\documentclass{article}

\usepackage{microtype}
\usepackage{graphicx}
\usepackage{subcaption}
\usepackage{multirow}
\usepackage{seqsplit} 
\usepackage{enumitem}
\usepackage{booktabs} 

\usepackage{hyperref}

\usepackage{tcolorbox}
\usepackage{listings}
\usepackage{courier}  
\usepackage{enumitem}

\usepackage[preprint]{neurips_2026}

\usepackage{amsmath}
\usepackage{amssymb}
\usepackage{mathtools}
\usepackage{amsthm}
\usepackage{hhline}

\usepackage[capitalize,noabbrev]{cleveref}

\theoremstyle{plain}

\theoremstyle{definition}

\theoremstyle{remark}

\usepackage{algorithm}
\usepackage{algorithmic}

\usepackage{tabularx}
\usepackage{booktabs}
\usepackage{ragged2e}
\usepackage{makecell}
\usepackage[normalem]{ulem} 

\usepackage[textsize=tiny]{todonotes}

\ifdefined\showcomments
  \newcommand{\chingan}[1]{\textcolor{blue}{(Ching-An: #1)}}
  \newcommand{\meghana}[1]{\textcolor{red}{(Meghana: #1)}}
  \newcommand{\allen}[1]{\textcolor{green}{(Allen: #1)}}  
\newcommand{\fanglei}[1]{\textcolor{orange}{(Fanglei: #1)}}    
\else
  \newcommand{\chingan}[1]{}
  \newcommand{\meghana}[1]{}
\newcommand{\allen}[1]{}
\newcommand{\fanglei}[1]{}
\fi

\title{RosettaSearch: Multi-Objective Inference-Time 
Search for Protein Sequence Design}

    \author{Meghana Kshirsagar\thanks{corresponding author}\\
    AI for Good, Microsoft\\
    Redmond\\
\texttt{meghana.kshirsagar@microsoft.com}\\
    \And
    Allen Nie\thanks{These authors contributed equally to this work.}\\
    Google DeepMind\thanks{prior work before joining} \\
    \And
    Ching-An Cheng\footnotemark[2]\\
    Google Research\thanks{partial work done previously at Microsoft Research}\\
    \And
    Fanglei Xue\footnotemark[2]\\
    Institute for Protein Design\\
    University of Washington\\
    \And
    Rahul Dodhia\\
    AI for Good, Microsoft\\
    Redmond\\    
    \And
    Juan Lavista Ferres\\
    AI for Good, Microsoft\\
    Redmond\\    
    \And
    Kevin K. Yang\\
    Microsoft Research\\
    New England\\
    \And
    Frank DiMaio\\
    Department of Biochemistry\\
    Institute for Protein Design\\
    University of Washington\\
}

\begin{document}

\maketitle

\begin{abstract}

We introduce RosettaSearch, an inference-time multi-objective optimization approach for backbone conditioned protein sequence design. We 
use large language models (LLMs) as a generative optimizer within a search algorithm capable of controlled exploration and exploitation, using rewards computed from RosettaFold3, a structure prediction model, under a strict computational budget. In a large-scale evaluation, we apply RosettaSearch to 400 suboptimal sequences generated by LigandMPNN (a state-of-the-art model trained for protein sequence 
design), recovering high-fidelity designs that 
LigandMPNN's single-pass decoding fails to produce. RosettaSearch's designs show improvements in structural fidelity metrics ranging between 18\% to 68\%, translating to a 2.5$\times$ improvement in design success rate.
We observe that these gains in success rate are robust 
when RosettaSearch-designed sequences are evaluated with an independent structure prediction 
oracle (Chai-1) and generalize across two distinct LLM 
families (o4-mini and Gemini-3), with performance scaling 
consistently with reasoning capability.

We further demonstrate that RosettaSearch improves the sequence fidelity of ProteinMPNN designs for \textit{de novo} backbones from the Dayhoff atlas, showing that the approach generalizes beyond native protein structures to 
computationally generated backbones.
We also demonstrate a multi-modal extension of RosettaSearch 
with vision-language models, where images of predicted protein structures are used as feedback to incorporate structural context to guide protein sequence generation. To our knowledge, this is the first large-scale demonstration that LLMs can serve as effective generative optimizers for backbone-conditioned protein sequence design, yielding systematic gains without any model retraining.




\end{abstract}

\chingan{for terminology, let's stick with inference-time search with LLM as generative optimizer, LLM optimizer, priority search vs sequential revision}

\section{Introduction}

Protein sequence design — the task of identifying amino acid sequences 
that fold into a target three-dimensional structure with desired functional 
properties - is a foundational challenge in computational biology with broad 
implications for drug discovery, enzyme engineering, the design of novel 
therapeutics etc. We focus specifically on \textit{backbone-conditioned sequence 
design}: given a fixed target backbone structure, find a sequence that 
confidently and accurately folds into it. We use the term \textit{fidelity} 
to refer to this property — whether a designed sequence, when its structure 
is predicted independently, recapitulates the target backbone with high 
confidence and geometric accuracy. Despite remarkable progress in deep 
learning approaches to backbone-conditioned sequence design such as 
LigandMPNN~\cite{dauparasLigandMPNN2023} and 
ProteinMPNN~\cite{dauparasProteinMPNN2022}, a persistent and practically 
significant gap remains: sequences generated by state-of-the-art models 
frequently exhibit low fidelity~\cite{anishchenko2021novo}. This matters 
because fidelity is the key bottleneck between computational protein design 
and experimental success: a sequence that cannot be reliably predicted to 
adopt the target structure is unlikely to do so in the wet lab, wasting 
costly synthesis and experimental validation effort.

The core difficulty is that backbone-conditioned sequence design is a high-dimensional, combinatorial optimization problem. The sequence space over amino acid alphabets is astronomically large, structural evaluation is expensive (requiring a full structure prediction call per candidate), and the relationship between sequence and fold is highly non-linear, with many local optima. 

Current approaches fall short in complementary ways. Single-pass sequence design models operate through a single forward pass of autoregressive decoding with no mechanism for self-correction: when a generated sequence lands in a low-fidelity region of sequence space, the model cannot recover since it cannot revisit or revise its output in response to downstream structural feedback. As a consequence, a meaningful fraction of designed sequences exhibit poor predicted structural confidence and 
geometric deviation from the target backbone, even when the model is given the 
native template as input. Optimization-based alternatives — reinforcement learning, directed evolution, and preference fine-tuning — enable iterative improvement but require task-specific training that is expensive and inflexible to new objectives. 

LLMs have been explored for protein design but only in one-shot settings, producing sequences with substantially lower structural fidelity than task-specific models~\cite{xia2025nature}. 
Generative optimization with LLMs has shown promise in other discrete optimization settings such as prompt, code, and hyperparameter 
optimization~\citep{khattab2023dspy, agrawal2025gepa}, but its potential as an optimizer for backbone-conditioned sequence design — 
capable of reasoning over structural feedback and proposing targeted sequence modifications — has not been systematically explored. See Section~\ref{sec:related:work} for a full discussion of related work.

We introduce \textbf{RosettaSearch} (See Figure \ref{fig:schematic}), an inference-time multi-objective optimization approach that addresses these limitations by deploying LLMs as generative optimizers within a structured search algorithm guided by structural rewards. Rather than asking an LLM to design a sequence from scratch — a task that strains its implicit knowledge of sequence-structure relationships — RosettaSearch instead asks it to \textit{refine} an existing candidate by reasoning over structured feedback from a structure prediction model (RosettaFold3~\cite{corley2025RF3}). This feedback combines global scalar rewards (derived from structure prediction metrics) with local, residue-level annotations identifying problematic regions. The LLM uses this feedback to propose targeted sequence edits within a \textit{priority search} algorithm that explores multiple modification trajectories in parallel, avoiding premature convergence. Because the reward function and feedback can be redefined at inference time, RosettaSearch requires no retraining or fine-tuning to adapt to new design objectives.


We also extend the framework to vision-language models (VLMs), incorporating rendered images of predicted structures as feedback to provide additional spatial context, and find that this produces richer reasoning in the model's chain-of-thought. To the best of our knowledge, this is the first large-scale demonstration that LLMs can function as effective generative optimizers for backbone-conditioned sequence design, yielding systematic gains over popular baselines without retraining or task-specific fine-tuning.

\begin{figure*}[h]
        \centering
        \includegraphics[width=0.9\linewidth]{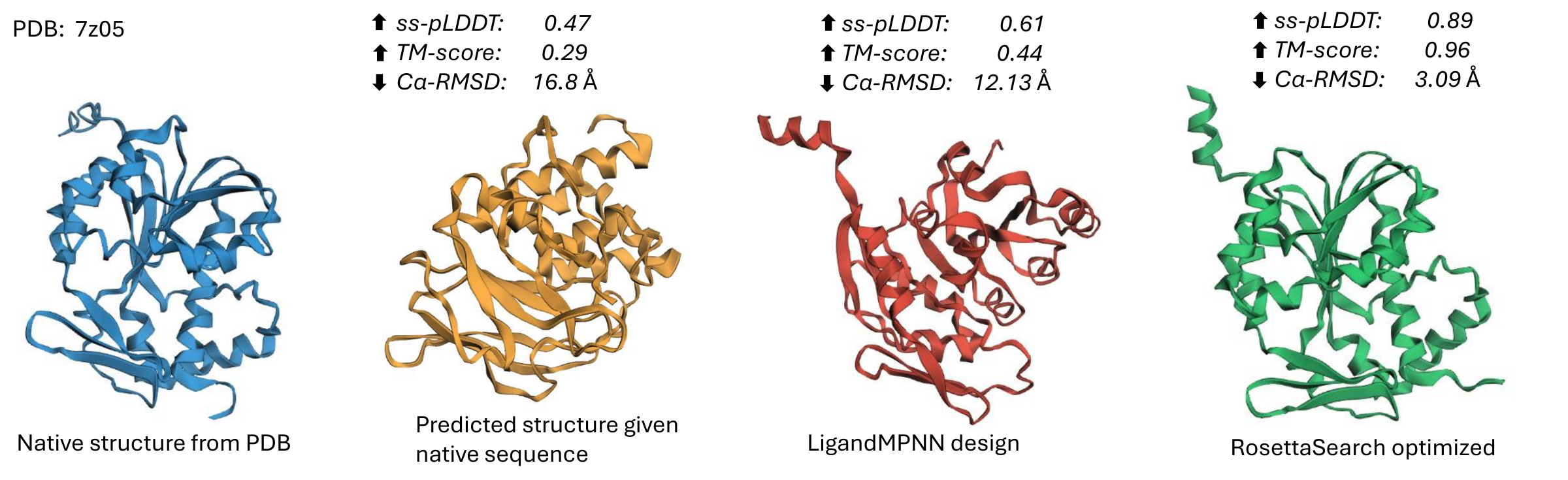}
        \caption{
        \textbf{Inference-time optimization with RosettaSearch dramatically  improves the structural fidelity of designed protein sequences.}
        For a protein in our dataset (PDB: \texttt{7z05}), RosettaSearch transforms a low-fidelity LigandMPNN design into a near-native structure, achieving large gains in pLDDT, TM-score, and RMSD using only inference-time feedback, without model retraining.
        We show the single sequence pLDDT obtained on each predicted structure (no multiple sequence alignment, i.e. MSA input). Also shown is the TM-score obtained by comparing the predicted structure to the PDB structure in blue. \textit{(1. Blue)} Native 3D structure downloaded from PDB. \textit{(2. Golden)} Predicted structure given the native sequence. \textit{(3. Red)} Predicted structure of the best design generated by LigandMPNN. \textit{(4. Green)} Predicted structure of the best design generated by RosettaSearch. Arrows next to the metrics indicate the desired direction of improvement. See Table \ref{tab:seqs} for the amino-acid sequences corresponding to each structure. Additional examples of structures that were optimized by RosettaSearch are shown in Figure \ref{fig:structure:res} (native monomers) and Figure \ref{fig:structure:res:dayhoff} (Dayhoff atlas backbones).  
        } 
    \label{fig:struc:res}
    \vspace{-3mm}
\end{figure*}

\section{Related Work}
\label{sec:related:work}
\paragraph{Backbone-conditioned sequence design.}
The dominant paradigm for backbone-conditioned sequence design involves 
learning a direct mapping from a fixed three-dimensional backbone structure (i.e. the 3D coordinates of backbone atoms are the inputs)
to a compatible amino acid sequence (amino acid sequence is the output). Models such as ProteinMPNN~\cite{dauparasProteinMPNN2022} 
and LigandMPNN~\cite{dauparasLigandMPNN2023} exemplify this approach, 
achieving strong sequence recovery rates on held-out structures and serving 
as widely adopted tools in the protein design community. However, these models 
operate through a single forward pass of autoregressive decoding and are 
optimized for sequence recovery rather than structural fidelity under 
independent evaluation. As a result, a meaningful fraction of their designs 
exhibit low confidence or poor geometric agreement with the target backbone 
when assessed by an independent structure predictor, and the models have no 
mechanism to detect or correct such failures at inference time. Our work 
addresses this gap by treating the output of such models as a starting point 
for iterative, feedback-driven refinement rather than a final answer.

\paragraph{Protein sequence optimization.}
Beyond single-pass design, a body of work frames protein sequence design as 
an iterative optimization problem. Energy-based 
methods~\cite{stracquadanio2011computational, zhou2024antigen} and 
reinforcement learning over mutation spaces~\cite{lutz2023top} can improve 
sequences toward target properties but typically require task-specific 
training or reward modeling. Directed evolution–style 
approaches~\cite{yang2019machine} offer more flexibility but rely on random 
or heuristic mutation operators without structured reasoning over structural 
feedback. Probabilistic~\cite{riesselman2018deep} and diffusion-based 
models~\cite{tang2025peptune} similarly require model retraining when 
objectives change. Preference learning approaches such as 
ResiDPO~\cite{xueResiDPO2025} fine-tune sequence design models using 
structural preference signals derived from AlphaFold, achieving improvements 
in structural fidelity but coupling the optimization to a fixed training 
procedure and dataset. In contrast, RosettaSearch performs optimization 
entirely at inference time, requiring no retraining or fine-tuning when 
objectives or design contexts change.

\paragraph{LLMs for protein design.}
Language models trained on protein sequences, such as 
ProtGPT2~\cite{ferruz2022protgpt2} and ESM-family models, can generate 
plausible \textit{de novo} sequences following the statistical patterns of 
natural proteins, but offer limited controllability for backbone-conditioned 
design tasks where a specific structural target must be matched. Frontier 
general-purpose LLMs such as GPT-4 have been evaluated on protein sequence 
generation in a one-shot setting, but produce sequences with substantially 
lower structural fidelity than task-specific models~\cite{xia2025nature}, 
reflecting their limited implicit knowledge of precise sequence-structure 
dependencies. The Virtual Lab~\cite{swanson2024virtual} and AI 
co-scientist~\cite{gottweis2025towards} explore multi-agent LLM workflows 
for protein design, integrating AI agents in a pipeline to propose and 
validate binders experimentally. While methodologically aligned in 
leveraging LLMs as components of a design workflow, these approaches focus 
on orchestrating multiple trained modules for a targeted task at small scale, 
rather than systematic inference-time optimization across hundreds of design 
problems. Our work differs fundamentally: we use the LLM not as a planner or 
coordinator, but as a generative optimizer that receives quantitative 
structural feedback and proposes targeted sequence edits within a principled 
search algorithm.

\paragraph{Generative optimization and inference-time search with LLMs.}
A growing body of work explores the use of LLMs as optimizers over discrete structured spaces, where textual feedback serves as a surrogate for 
gradients~\citep{khattab2023dspy, cheng2024trace, yuksekgonul2024textgrad, 
nie2024importance}. Inference-time search techniques have been incorporated to enhance this process, including beam search~\citep{sun2023allies, 
pryzant2023automatic, chen2024prompt}, Monte Carlo Tree Search~\citep{wang2023promptagent}, Gibbs 
sampling~\citep{xu2023reprompting}, FunSearch~\citep{romera2024mathematical}, 
and GEPA~\citep{agrawal2025gepa}. These methods have been applied to optimizing prompts, code, and hyperparameters, and more recently to 
scientific discovery tasks including system code 
optimization~\citep{wei2024improving}, automated scientific 
discovery~\citep{ghareeb2025robin}, and chemistry 
reasoning~\citep{narayanan2025training}. Our work extends this line of research to backbone-conditioned protein sequence design, a setting that 
poses distinct challenges: the search space is combinatorial over a structured biological alphabet, evaluation via structure prediction is 
expensive, and the optimization must balance multiple structural objectives simultaneously while satisfying biological validity constraints. Our work is inspired by the Trace framework~\citep{cheng2024trace}, which we extend with a priority-based search algorithm, multi-objective reward formulation, and residue-level textual feedback specifically designed for the protein sequence design setting.

\section{Design of RosettaSearch}
\vspace{-1mm}

RosettaSearch adopts generative optimization~\citep{khattab2023dspy,cheng2024trace} to improve protein sequence design. Generative optimization employs generative models (such as LLMs and VLMs) as optimizers to analyze and directly propose changes to parameters using rich feedback, such as a combination of scores and texts. 
Below we discuss metrics important to protein design, the design of reward and feedback to the generative optimizer, and how we coordinate generative optimizer calls to produce candidates that improves the starting protein sequences.

\begin{figure}
    \centering
    \includegraphics[width=\linewidth]{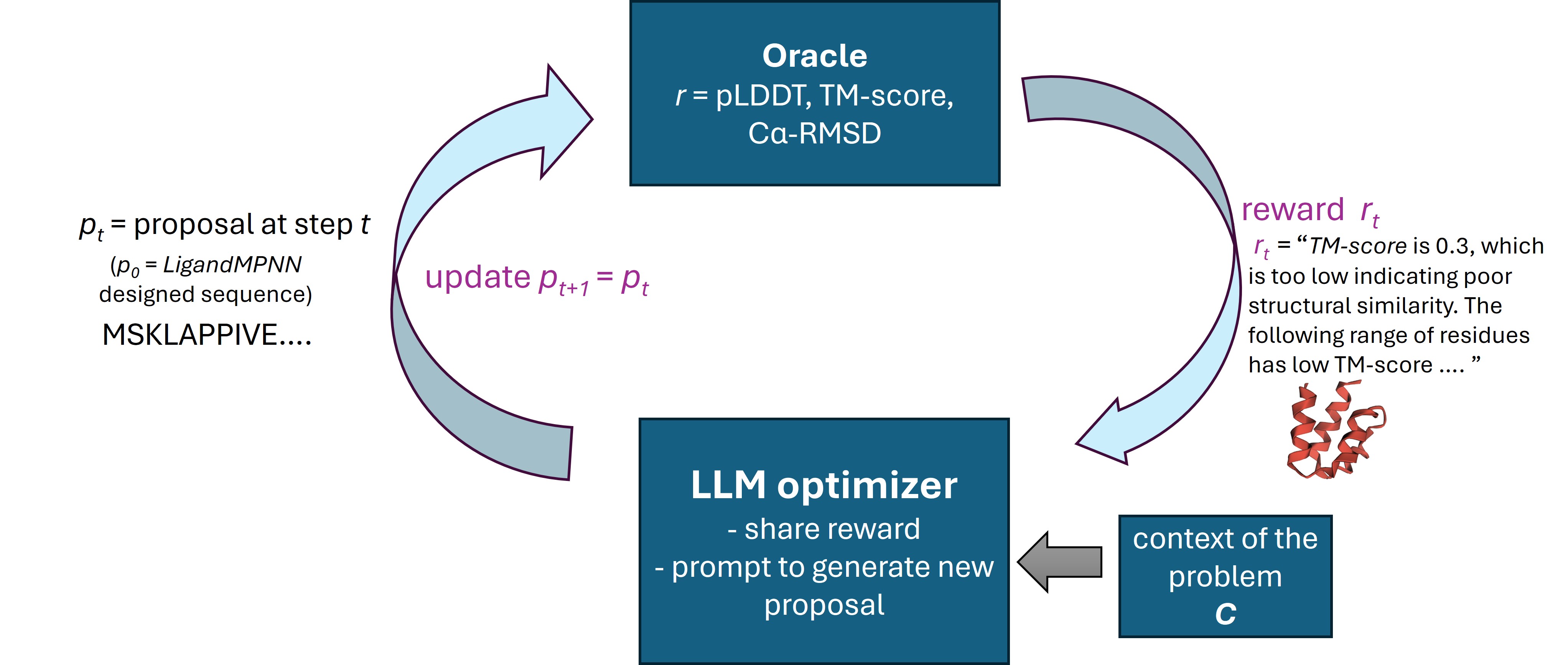}
    \caption{Schematic showing the approach followed by RosettaSearch}
    \label{fig:schematic}
\end{figure}

A \textit{reference} native structure or \textit{backbone} structure denotes the target 3D coordinates that define the desired fold, providing a fixed structural template against which candidate sequences (upon obtaining their predicted structure) are evaluated for fold consistency, structural accuracy etc.

\noindent\textbf{Notation:} We use $s \in A^L$ to denote the proposed protein amino-acid sequence, and $s_t \in A^L$ to denote the proposal at step $t$ within a generation trajectory, where $A$ is the amino-acid alphabet and $L$ is the sequence length.  We use $\hat{\mathbf{x}} = f(s)$ to denote the predicted protein structure, where $\hat{\mathbf{x}}$ contains the 3D atomic coordinates and $f$ is the protein structure prediction model, such as RosettaFold3. We use $s^* \in A^L$ to denote a reference sequence, if available, such as the native protein sequence whose structure we want to match. We use $\mathbf{x}^*$ to denote the native structure's backbone 3D atomic coordinates.

\vspace{-1mm}
\subsection{Metrics used to quantify structural fidelity of designed protein sequences}
\vspace{-1mm}
Protein sequences fold to form protein structures and structure determines function. For example, in enzymes like lysozyme, catalytic activity depends on the precise three-dimensional arrangement of the active-site residues.
Likewise, therapeutic design for COVID must account for the fact that binding between the SARS‑CoV‑2 spike protein and the human ACE2 receptor is governed by precise three‑dimensional shape complementarity at their interface.

Hence, while designing proteins with a desired function, it is essential that the generated protein sequence adheres to the given desired structure. Metrics that measure such adherence can be considered \textit{fidelity metrics}.
We calculate the following metrics, defined in detail below, for the listed reasons: 

\noindent\textbf{pLDDT}: predicted Local Distance Difference Test (pLDDT) \cite{tunyasuvunakool2021highly} of a protein sequence's predicted structure, denoted as $pLDDT(\mathbf{\hat{x}})$, is a per-residue (i.e. per position of the protein sequence) confidence metric output by biomolecular structure prediction models such as RosettaFold3~\cite{corley2025RF3} and AlphaFold3~\cite{abramsonAF32024}. It estimates the expected local accuracy of the predicted atomic positions with values $\in [0, 100]$. Higher pLDDT values ($>$70) indicate greater confidence in local structural geometry of the molecule, while lower values highlight regions likely to be disordered or incorrectly modeled. 
pLDDT has been shown to correlate with local structural accuracy and is widely used as a proxy for predicted fold reliability, or how likely the sequence is to adopt the predicted fold once synthesized in the wet-lab \cite{yin2022benchmarking}. 

C$\alpha$-RMSD and TM-score are widely used as a measure of the consistency of a designed protein with respect to the desired structure / template. Sequences whose predicted structures have poor C$\alpha$-RMSD and TM-score may not exhibit the desirable behaviour upon synthesis.

\noindent\textbf{C$\alpha$-RMSD}: This metric compares the corresponding backbone $\alpha$-carbon atoms of the structures of two biomolecules. It calculates the root mean square deviation (RMSD) between them, and requires that the underlying sequences of the two structures being compared be of equal length. Details of the implementation are provided in the Appendix in Section \ref{appendix:calpha}.

\noindent\textbf{TM-score}: Template-matching score (TM-score) is used as a primary measure of global structural similarity between the predicted structure $\hat{\mathbf{x}}$ and the reference $\mathbf{x}^*$, denoted as TM-score$(\mathbf{x}, \mathbf{x}^*)$. Unlike C$\alpha$-RMSD, TM-score is length-normalized (which allows $L \neq L^*$) and less sensitive to local deviations, making it particularly suitable for comparing structures of similar but not identical sequences.

\subsection{Formulating effective reward and feedback} \label{sec:reward and feedback design}

We discuss the formulation of reward and text/multimodal feedback for the generative optimizer to improve protein designs. Overall the feedback contains information about a protein design's global and local quality, as well as constraint violation. An example feedback is shown in Figure \ref{fig:prompt} (bottom). 

\vspace{-1mm}
\subsubsection{Reward and Feedback}
\vspace{-1mm}
We define a structured reward and feedback that give global and local information about a protein candidate. The feedback type is configurable for each run.

\vspace{-1mm}
\paragraph{Global Reward}
Given a proposed protein sequence, we first predict its three-dimensional structure using RosettaFold3 (RF3 produces multiple predictions, of which we pick the best one -- the one with the highest average pLDDT). 
Then we compute every metric that we are optimizing by averaging over the per-residue (i.e. per position) metric values \chingan{what does per-residue mean}and scaling the result to lie within [0,1]. $C\alpha$-RMSD values exceeding a predefined penalty threshold incur an explicit negative correction, to indicate a sharp degradation in structural fidelity beyond acceptable deviations.\chingan{what does this mean exactly. From the writing so far, it's not clear how the reward is computed}. In addition to the reward value itself, textual description of the quality of the value is included (for example: ``0.3 is a very low reward. Significant improvements needed.''). These signals provide global information about the objective quality of a candidate sequence.

\vspace{-1mm}
\paragraph{Local Text Feedback}
We found that optimization using global information alone do not result in substantial improvements to the metrics in our preliminary experiments. To address this, we additionally introduce \textit{local} text feedback to help direct the LLM's optimization towards problematic regions. For pLDDT, low-confidence regions are detected via adaptive pLDDT thresholds and reported as contiguous residue ranges, encouraging localized sequence refinement rather than global rewrites. Analogously, with $C\alpha$-RMSD, per-residue inter-atomic distances between $\mathbf{x}$ and $\mathbf{x}^*$ are binned, and contiguous residue intervals corresponding to each distance bin are reported. A similar text feedback is constructed for TM-score. 

\vspace{-1mm}
\paragraph{Multi-modal Learning through Image Feedback}
When using vision-language models as the optimizer, we extend our approach to use also a rendered image of the protein’s 3D structure as feedback. 
The structural image provides additional global spatial context about the protein template, including secondary structure organization, domain topology, and exposed versus buried regions, that are difficult to convey through sequence alone. 

\vspace{-1mm}
\subsubsection{Feedback to Prevent Reward Hacking}
When optimizing for a scalar reward, 
we notice that LLMs often discover strategies that maximize the objective while deviating from the intended notion of valid protein design. 
In particular, proxy rewards such as pLDDT become increasingly decoupled from true design quality under optimization pressure. The model exploits this mismatch by repeatedly selecting amino acids with high metrics, replicating short motifs known to yield consistently high pLDDT, producing trivially short or excessively long sequences that inflate the reward, appending long runs of structurally benign residues (e.g., poly-alanine). In cases involving native enzyme redesign where a native protein sequence is available as additional context information, the LLM returns the native protein sequence itself as a local optimum. These behaviors resemble classic reward-hacking failure modes~\citep{skalse2022defining}, where the proxy objective ceases to faithfully represent the underlying goal. 

Inspired by constrained optimization, we introduce text feedback to restrict the solution space and restore a measure of alignment between the optimized reward and genuine design intent.
These constraints are presented as text feedback given to the LLM during optimization to address the observed failure modes above.
While these signals do not directly modify the reward value, they affect the LLM’s subsequent proposals and effectively act as soft constraints.


\noindent\textbf{Constraint to avoid repetitive regions}: If the number of repeated 6-mers is greater than 5\% of the total number of 6-mers in a proposed protein sequence, the feedback will indicate that the sequence is non-ideal: ``the generated sequence is too repetitive''. Otherwise, the feedback says: ``the generated sequence is sufficiently non-redundant''.

\noindent\textbf{Constraint on length}: If the generated sequence's length exceeds (or trails) the expected length, we add the feedback ``the generated sequence is longer (shorter) than the desired length of $n$'' to the feedback.

\noindent\textbf{Retaining ligand binding}:
To preserve functional interactions during optimization, we encode ligand-binding constraints by providing the model with residue-level annotations identifying positions involved in ligand binding. 

\vspace{-1mm}
\subsection{Generative optimization with LLMs}
\vspace{-1mm}

We consider two ways to deploy generative optimizers: Sequential Revision and Priority Search.

\vspace{-1mm}
\subsubsection{Iteration Schemes}

\vspace{-1mm}
\paragraph{Sequential Revision}
The first version is sequential revision which iteratively calls the generative model to improve the latest protein candidate at hand using feedback. 
Our work adapts the LLM optimizer in Trace~\citep{cheng2024trace} (a ReAct style optimization agent) and treats the protein sequence as a trainable text variable. At each iteration, the optimizer constructs a structured problem instance from intermediate states, outputs, and scalar or textual feedback consisting of variables, constraints, code, and observed outcomes. This structured representation is provided to an LLM, which reasons about how changes to the trainable variables should follow the feedback signal and proposes targeted updates. We run the optimizer iteratively to improve the starting protein sequence. 

\noindent{\textbf{Priority Search}}
Iterative sequential optimization is prone to early convergence to local optimum and optimization instability. 
As an alternate strategy, we design a priority-based parallel search algorithm. 
Analogous to sequential revision, we treat the protein sequences as a text parameter. Our search algorithm maintains a buffer storing all the previous design attempts and their evaluation results; the buffer is initialized with the starting protein sequence. In each iteration, we take the top-$K$ performing candidates from the buffer and evaluate them in parallel to obtain the reward and feedback described above.
%
Then we generate generate $K$ new candidates in parallel based on feedback given these top $K$ candidates (one new candidate is independently generated for each candidate). These new candidates are then evaluated using the feedback oracle again, and all 2$K$ candidates with their latest statistics are added back to the buffer. 
During the optimization, the buffer contains all parameter proposals and tracks all scores received so far. 
%
See Algorithm \ref{alg:priority-search} for an overview. 

\subsubsection{Multiobjective Optimization}
\chingan{I don't know how to interpret this section with the information given in the previous sections. I'm confused by what the actual overall algorithm is. Which version was used at the end to produce the experiments? -- DONE }

We explored several strategies for handling the inherently multiobjective nature of protein sequence optimization, where structural confidence, geometric fidelity, and sequence validity must be balanced. We describe the other variants that did not work as well in Appendix Section \ref{appendix:multiobj}. 

We tried a weighted-sum formulation that combines all objectives into a single scalar reward using fixed coefficients: $  R(s) = w_{tm} \textrm{TM}(\mathbf{x},\mathbf{x}^*) + w_p\;  \textrm{pLDDT}(\mathbf{\hat{x}}) + w_r \; C\alpha\textrm{-RMSD}(\mathbf{x},\mathbf{x}^*)$
 While each approach exposes different trade-offs between stability and flexibility, the weighted-sum formulation provided the most consistent optimization behavior in our setting and was therefore the one adopted for the final multiobjective configuration.

\subsubsection{Meta Design}

\paragraph{Constraining exploration using step-size}
To regulate the magnitude of sequence modifications during optimization, we introduce a textual step-size mechanism that constrains the number of residue edits permitted at each optimization step. 
We encode step-size directly in the natural-language objective provided to the LLM, effectively controlling the rate of exploration in sequence space. Concretely, we define five regimes: (i) unconstrained, allowing arbitrary modifications; (ii) conservative, permitting one to two residue changes per step; (iii) moderate, allowing up to five residue changes; (iv) aggressive, allowing up to ten residue changes; and (v) drastic, allowing up to twenty residue changes per step. This design mirrors classical step-size selection in gradient-based optimization, where smaller steps encourage local refinement and stability, while larger steps enable broader exploration at the risk of overshooting.

\noindent{\textbf{Best-of-N}}
For each protein design task, we adopt a best-of-N selection strategy to dampen the effects of the hyper-parameters as well as the lack of strict monotonicity in the optimization. We run RosettaSearch under three distinct textual step-size regimes, each with a fixed evaluation budget of number of evaluations = 25 structure evaluations. This yields a total of $N = 3 \times \textrm{num\_evals}$ candidate sequences per protein. All runs share the same reward and  feedback on quality and constraint, differing only in the magnitude of sequence edits permitted at each step. The final output sequence is selected as the highest-reward candidate among all $N$ evaluated sequences. 
\chingan{what is the reward criterion? I thought it is multi objective}
This is analogous to best-of-$N$ sampling in LLM decoding, where multiple candidate generations are evaluated and the highest-scoring output is selected to improve reliability under a fixed compute budget. 
This query-limited strategy places our method closer to budgeted black-box optimization than large-scale reinforcement learning, and contrasts with approaches that require thousands of gradient updates for model training.

\vspace{-1.5mm}
\subsection{Final design choices}
\vspace{-1mm}
We use \texttt{o4-mini} as the LLM. The multi-objective weight parameters that worked optimally were: $w_p=\frac{1}{11}$, $w_{tm}=\frac{5}{11}$ and $w_{r} = \frac{5}{11}$. On step-size, we found `aggressive' (10 amino acids modified), `moderate' (5 amino acids) and `unconstrained' (any edits) to be the three optimal settings that offered the best exploration / exploitation behavior. For each optimization task, we run one Priority Search for each one of the three optimal step-sizes to get 3 sets of trajectories with the limit on number of evaluations set at 75, allowing 25 evaluations per step-size. 
Consequently, the number of candidates explored in each round is 3 and the number of proposals per candidate are 2.



\vspace{-1mm}
\section{Datasets}
\vspace{-1mm}
\chingan{Give an overview and the goall take about RossettaFold3 is used as the evaluator}

\chingan{Not sure if putting dataset section early is common. Another place to put it is in the experiment setup later after introducing the method.--DONE}

\paragraph{LigandMPNN dataset consisting of PDB Monomers}
To evaluate the LLM's capability in structure-preserving sequence optimization, we curated a dataset from PDB-D \cite{xueResiDPO2025}. We randomly sampled $\approx$400 monomeric structures. For each structure, we utilize both the native sequence and a set of redesigned sequences generated via LigandMPNN~\cite{dauparasLigandMPNN2023}. This paired data allows us to assess whether the LLM can optimize a given protein sequence to improve fidelity relevant metrics (e.g., folding confidence) while strictly maintaining the original structure.

\vspace{-1mm}
\paragraph{The Dayhoff dataset}
The Dayhoff BackboneRef dataset~\cite{yang2025dayhoff} consists of \textit{de novo} protein backbones generated with RFDiffusion~\cite{watsonRFDiffusion2023}, where sequences are designed using ProteinMPNN~\cite{dauparasProteinMPNN2022}. These structures exhibit high baseline fidelity, providing a stringent testbed to evaluate whether optimization methods can improve well-folded designs without overfitting to a sequence design pipeline. 

\noindent\textbf{BindCraft Sequences}
To assess whether RosettaSearch can improve fidelity beyond \textit{de novo} sequence generation, we evaluate our method on 50 protein binders randomly selected from the dataset released by the BindCraft study~\cite{pacesaBindCraft2025}. These binders were originally designed using a computationally intensive pipeline involving large-scale structure prediction and optimization, and represent high-quality candidates with strong predicted binding affinities to their target enzymes.
This setting provides a stringent test of whether inference-time search can refine already optimized designs without access to the original training or fine-tuning procedures used by BindCraft.

\noindent{\textbf{Remark}} 
Note that when using RosettaFold3~\cite{corley2025RF3} as the structural oracle, we evaluate single-sequence pLDDT without multiple sequence alignment (MSA) conditioning to avoid inflating confidence estimates through external evolutionary priors\chingan{what does external evolutionary prior mean}. MSA inputs provide strong homology-derived signals that can dominate structure confidence predictions, even for sequences whose foldability is weak in isolation, as found in prior work \cite{korbeld2025limitations}. In a protein redesign setting—where candidate sequences may deviate substantially from natural families, MSA-based evaluation can therefore overestimate fidelity by leveraging information not intrinsic to the proposed sequence. Computing pLDDT in single-sequence mode yields a stricter, sequence-intrinsic confidence measure that better aligns with the optimization objective and avoids rewarding designs solely due to evolutionary context rather than structural feasibility.

\vspace{-1mm}
\section{Experimental Results}
\vspace{-1mm}

We conduct a comprehensive empirical evaluation of RosettaSearch across multiple protein sequence optimization settings. Our experiments are designed to answer the following questions: (i) whether inference-time, feedback-driven optimization can reliably improve fidelity metrics over various baselines and to what extent; (ii) whether gains in performance persist under independent structure predictors and diverse initialization regimes; (iii) understanding the role that context plays in improving the objectives; (iv) understanding whether visual information through images of protein structures is helpful in generating sequences; (v) whether the LLM reasoning carries useful information.

\chingan{Describe an overview of the experiments. What are we trying to study? what are the conclusions?  We need some overview to help the reader navigate through this long section.}


\chingan{Need to describe the experiment setup first before jumping into baselines, metrics etc..  We need to know the goal of the experiments first so that we can know whether they make sense or not.}


\chingan{mention llm used in the experiments}

\vspace{-1mm}
\subsection{Evaluation metrics}
\label{sec:eval:metrics}
\vspace{-1mm}

We measure the improvements in fidelity metrics (pLDDT, TM-score, C$\alpha$-RMSD) obtained by our approach. We evaluate the resulting sequences on additional criteria below, since pLDDT, TM-score and C$\alpha$-RMSD could have been used during optimization,

\noindent\textbf{Success rate-1}: We define design \textit{success rate} as the fraction of protein sequences whose predicted structures simultaneously achieve high confidence and high structural similarity to the reference. Specifically, a design is considered successful if its predicted structure attains a pLDDT of at least 0.8 and a TM-score of at least 0.8 when compared to the reference structure. The success rate is computed as the proportion of sequences in the dataset satisfying both criteria.

\noindent\textbf{Success rate-2}: We define another success rate, which uses all three objectives and sets a more stringent criterion for success given by pLDDT$\geq 0.8$, TM-score$\geq 0.8$ and C$\alpha$-RMSD $\leq$ 1.5.


\begin{table}[ht]
\centering
\setlength{\tabcolsep}{0.7pt} 
\renewcommand{\arraystretch}{1.2} 
\caption{Sequential revision vs Priority search, absolute and \% improvement (or reduction). For absolute improvement, the average value of the metric before and after are shown. See Appendix Table \ref{table:stat:sig} for statistical significance.}
\label{table:iter:rollout}
\begin{small}
\begin{tabular}{p{2.6cm}p{1.9cm}p{1.6cm}p{1.9cm}}
\toprule
\multicolumn{4}{l}{\textit{Single objective} (pLDDT)} \\ \hline
& ss-pLDDT   & TM-score   & $C\alpha$-RMSD   \\ 
\cline{2-4} 
Sequential Rev. & 0.59$\rightarrow$0.67 & 0.44$\rightarrow$0.54 & 19.2$\rightarrow$14.6 \\
& 17.0\% & 35.5\% & -20.0\% \\ 
Priority Search & 0.59$\rightarrow$0.69 & 0.44$\rightarrow$0.57 & 19.2$\rightarrow$13.0 \\
& \textbf{19.0}\% & \textbf{50.9}\% & \textbf{-30.1}\%
\\ \hline
\\
\multicolumn{4}{l}{\textit{Multi objective} (ss-pLDDT, TM-score, $C\alpha$-RMSD)} \\ \hline
& ss-pLDDT  & TM-score  & $C\alpha$-RMSD  \\ 
\cline{2-4} 
Sequential Rev.  & 0.59$\rightarrow$0.71 & 0.44$\rightarrow$0.61 & 19.2$\rightarrow$11.7 \\
& \textbf{24.2}\% &  65.1\% &  -32.6\% \\ 
Priority Search & 0.59$\rightarrow$0.68 & 0.44$\rightarrow$0.62 & 19.2$\rightarrow$11.2 \\
& 18.0\% \; & \textbf{67.7}\% \; & \textbf{-37.7}\% \\ 
\bottomrule
\end{tabular}
\end{small}
\end{table}

\noindent\textbf{Evaluation using a different structure prediction model}:\\
Evaluating designed sequences using the same structure prediction model that provides feedback during optimization can lead to inflated performance estimates due to overfitting to model-specific inductive biases. In our framework, RosettaFold3 (RF3) is used during the optimization loop to assess fidelity and guide sequence edits; therefore, success rates computed solely using RF3 may overstate generalization. To mitigate this effect, we additionally evaluate all final designs using another independent structure prediction models, Chai-1 \cite{discoveryChai-12024} 
which differs in architecture, training data, and inductive biases. 

\vspace{-1mm}
\subsection{Random mutations baseline}
\vspace{-1mm}
To isolate the contribution of search from improvements due purely to random exploration, we introduce a baseline that is: (1) compute-matched — given the same oracle budget as RosettaSearch, (2) information-matched — provided with the same residue-level feedback identifying low-pLDDT and low-TM-score regions, (3) initialization-matched -- it is also initialized with LigandMPNN designs. Further, at each step, it applies the same number of residue modifications as RosettaSearch, but makes random amino acid substitutions within the low-metric regions. 
The computational budget is controlled by allowing this baseline the same number of oracle evaluations, measured as the number of calls to RosettaFold3, as used by our method.
The best random mutation design is selected by best-of-N sampling, picking the highest scoring one across runs.

As seen from the results in Table \ref{tab:success:rate}, this baseline worsens the success rate of LigandMPNN designs, despite the guidance and focused feedback from a powerful oracle like RosettaFold3. We can thus conclude that the LLM's ability to interpret that feedback and make targeted, chemically informed substitutions is the critical ingredient.

The baseline effectively fails to improve over its LigandMPNN initialization -- the highest scoring candidate across runs is typically the original LigandMPNN design itself, 
indicating that random substitutions within flagged regions consistently degrade rather than improve structural fidelity. This result demonstrates that residue-level feedback alone is not sufficient -- the LLM's ability to interpret that feedback and make 
targeted, chemically informed substitutions is the critical ingredient driving RosettaSearch's 
improvement in success rate.


\subsection{Cumulative results} 
We combine optimization trajectories from both single-objective (pLDDT only) and multi-objective (pLDDT, TM-score with and without C$\alpha$-RMSD) runs, then select the best-performing sequence across all generated candidates. This approach leverages the complementary strengths of different optimization strategies -- single-objective runs may find sequences that excel in confidence prediction, while multi-objective runs balance multiple structural fidelity criteria. As expected, cumulative selection yields the strongest overall results, albeit with the combined budget of all the runs.

\subsection{Priority search generally outperforms sequential revision}
\vspace{-1mm}
We observe that a priority-based parallel search strategy consistently outperforms purely sequential revision for protein sequence refinement as seen in Table \ref{table:iter:rollout}. In sequential revision, each step greedily conditions on the current best sequence, which can lead to premature convergence and error accumulation, particularly when early edits introduce subtle structural inconsistencies that later steps cannot recover from. In contrast, priority search explores multiple candidate modification trajectories in parallel by sampling diverse sequence proposals at each step and evaluating them independently. This allows the method to defer commitment to a single trajectory until downstream structural signals (e.g., pLDDT and TM-score) are observed, improving robustness to noisy intermediate evaluations. By selecting the best-performing sequence from a pool of proposals rather than relying on a single optimization path, priority search better captures long-horizon effects of residue substitutions and yields higher overall success rates under identical query budgets.



\begin{table*}[t]
\centering
\caption{\textit{\textbf{(Upper section)}} shows baseline success rates for various sets of sequences used as starting points for optimization as well as the performance of a randomized mutations baseline that is also guided by RosettaFold3 (RF3). The success rate columns shows how well the best designs do when evaluated by RF3 and Chai-1 respectively. Here, for each protein 
redesign, we take the best design produced by our approach and re-evaluate it 
using Chai-1, an independent structure prediction model with different 
architecture and training data from RF3. \textit{\textbf{(Lower section)}} Design success rates obtained by RosettaSearch under different initialization and optimization 
settings. To assess the RosettaSearch performance, compare each initialization method (a), (b) in the lower section to its corresponding baseline in the upper section.}
\begin{small}
\label{tab:success:rate}
\renewcommand{\arraystretch}{1.1}
\begin{tabular}{p{3.7cm}p{6cm}p{1.1cm}p{1.2cm}}
\toprule
& & \textbf{Success Rate (RF3)} & \textbf{Success Rate (Chai-1)} \\
\midrule
\multicolumn{4}{l}{\textit{Baseline success rates of various starting sequences}} \\
\midrule
\multicolumn{2}{l}{(a) Native monomer sequences}  & 7.0\% & 2.4\% \\
\multicolumn{2}{l}{(b) LigandMPNN designs } & 7.9\% & 3.1\% \\
\multicolumn{2}{l}{(c) Random protein sequences } & 0.0\%    & 0.0\%  \\\\
\multicolumn{2}{l}{Random mutations baseline initialized with LigandMPNN designed sequences} & 7.9\% & 3.1\% \\
\multicolumn{2}{l}{\; \; (shows no improvement over starting sequences' success rate)} & & \\
\midrule
\textit{RosettaSearch} & & & \\
\textbf{Initialization} & \textbf{Objective function} & & \\
\midrule
\multirow{4}{*}{(a) Native monomer sequences}
 & Single obj.\ (ss-pLDDT)                          & 7.5\%  & 4.2\% \\
 & Multi-obj.\ (ss-pLDDT, TM-score)                 & \textbf{9.1\%}  & \textbf{4.7\%} \\
 & Multi-obj.\ (ss-pLDDT, TM-score, C$\alpha$-RMSD) & 8.4\%  & 4.3\% \\
 & Cumulative                                        & \textbf{10.1\%} & — \\

\midrule

\multirow{4}{*}{(b) LigandMPNN designs}
 & Single obj.\ (ss-pLDDT)                          & 18.4\% & 5.0\% \\
 & Multi-obj.\ (ss-pLDDT, TM-score)                 & 17.1\% & \textbf{6.7\%} \\
 & Multi-obj.\ (ss-pLDDT, TM-score, C$\alpha$-RMSD) & \textbf{20.5\%} & 5.3\% \\
 & Cumulative                                        & \textbf{21.6\%} & — \\



\bottomrule
\end{tabular}
\end{small}
\end{table*}

\vspace{-1mm}
\subsection{RosettaSearch improves sequence fidelity}
\vspace{-1mm}
On a large dataset of $\approx$400 protein redesign examples involving monomer protein structures with low single-sequence fidelity metrics, using single objective optimization we find that RosettaSearch improves single sequence pLDDT (i.e. ss-pLDDT) by 18.1\%, TM-score by 56.9\% and reduces $C\alpha$-RMSD by 31.2\% as seen in Table \ref{table:iter:rollout}. Note that the improvements are much higher when there are no constraints on the generation; for instance, ss-pLDDT improves by 57\% (see Appendix Figure 3(a)). We find that multi-objective optimization has a bigger improvement in metrics overall.

\begin{table}[h]
\centering
\renewcommand{\arraystretch}{1.1}
\setlength{\tabcolsep}{5pt}
\caption{Performance comparison with a more stringent success rate criterion. RosettaSearch results are shown with multi-objective optimization with all three objectives. \meghana{TODO: report statistical significance}} 
\label{table:success:rate2}
\begin{tabular}{lcc}
\toprule
 & LigandMPNN & RosettaSearch \\
\midrule
success rate-2 & 2.5\% & 8.9\% \\ 
number of successes & (10/394) & (35/394) \\ 
 \bottomrule
\end{tabular}
\end{table}


We further investigate whether these improvements translate to increase in design success rates in three settings involving different starter protein sequence distributions as shown in Table \ref{tab:success:rate}: (a) initializing with a suboptimal native sequence (b) initializing with candidates produced by a popular backbone conditioned sequence generation approach, LigandMPNN. We find that RosettaSearch achieves a similar design success rate as LigandMPNN in setting (a), where LigandMPNN directly conditions on the input template structure, while our approach uses this information indirectly through feedback. Further, starting with the sequences designed by LigandMPNN in setting (b), RosettaSearch achieves a 2.5x improvement (to 20.53\%) over the design success rate of LigandMPNN (7.95\%). This equates to going from 32 successful designs to 79 successful designs.

Finally, when starting from random protein sequences, we find that RosettaSearch is able to achieve a design success rate of 8.1\% that is equal to that of LigandMPNN. This rate is achieved in the setting where native monomer sequences are available as contextual information, to the LLM optimizer.

When the metrics are obtained using an alternate `oracle', Chai-1~\cite{discoveryChai-12024} (please refer to Section \ref{sec:eval:metrics} for why we do this), the native sequences as well as all methods exhibit substantially lower absolute structural metrics as seen in Table \ref{tab:success:rate} (last column), partly due to Chai-1's known tendency to produce conservative predictions as compared to RosettaFold3 (See Figure \ref{fig:chai:vs:rf3} for sequence-level predicted pLDDTs and TM-scores that show the lower metrics). 
Importantly though, the relative improvements remain consistent: our method achieves approximately a 2× increase in success rate over baselines under both RF3 and Chai-1 evaluation (an increase from 2.4\% to 4.7\% as well as an increase from 3.1\% to 6.7\%). 


To assess the generality of RosettaSearch across LLM families, we evaluate two additional models under identical experimental conditions. Gemini-3, from a completely different model family than o4-mini, achieves a design success rate of 19.4\%, comparable to o4-mini's 20.5\%, 
demonstrating that the gains are not specific to any single model family. Within the same family, o3-mini achieves a success rate of 12.3\%, substantially above the 7.9\% 
LigandMPNN baseline but below o4-mini, consistent with its weaker reasoning capability. Taken together, these results suggest that performance scales with reasoning capability across and within model families, and that any sufficiently capable reasoning model can serve as 
an effective generative optimizer within the RosettaSearch framework. This is in line with findings from prior work~\cite{greisen2025reasoning}, who independently demonstrate that reasoning 
capability is the critical factor distinguishing successful from unsuccessful LLMs on protein design tasks.

We next look at the diversity of the generated sequences. Figure \ref{fig:seqid} shows the distribution of per-protein average sequence identity between the starting sequences and the two best generated sequences on the LigandMPNN dataset. Although the mean sequence identity is high (82.4\%), the distribution is wide, ranging from near-identical sequences to substantial divergence. This indicates that the method adapts the extent of sequence modification on a per-protein basis, making conservative edits when sufficient and exploring more aggressive redesigns when required. Results on whether the LLM prefers certain parts of the sequence to mutate are in the Appendix in Section \ref{appendix:updates}.
In the last row of Table \ref{tab:success:rate}, results from multi-modal sequential revision of the LigandMPNN protein sequences are shown, where visual feedback is provided at each iteration.
 
\textbf{Computational cost:} The wall clock time per protein depends on the length of the protein and ranges from $\approx$2 min to $\approx$1.4 hours, with an average of 30.6 minutes and is dominated by RosettaFold3 inference (over the PDB dataset which has a total of 400 proteins).

\begin{figure*}
    \includegraphics[width=\linewidth]{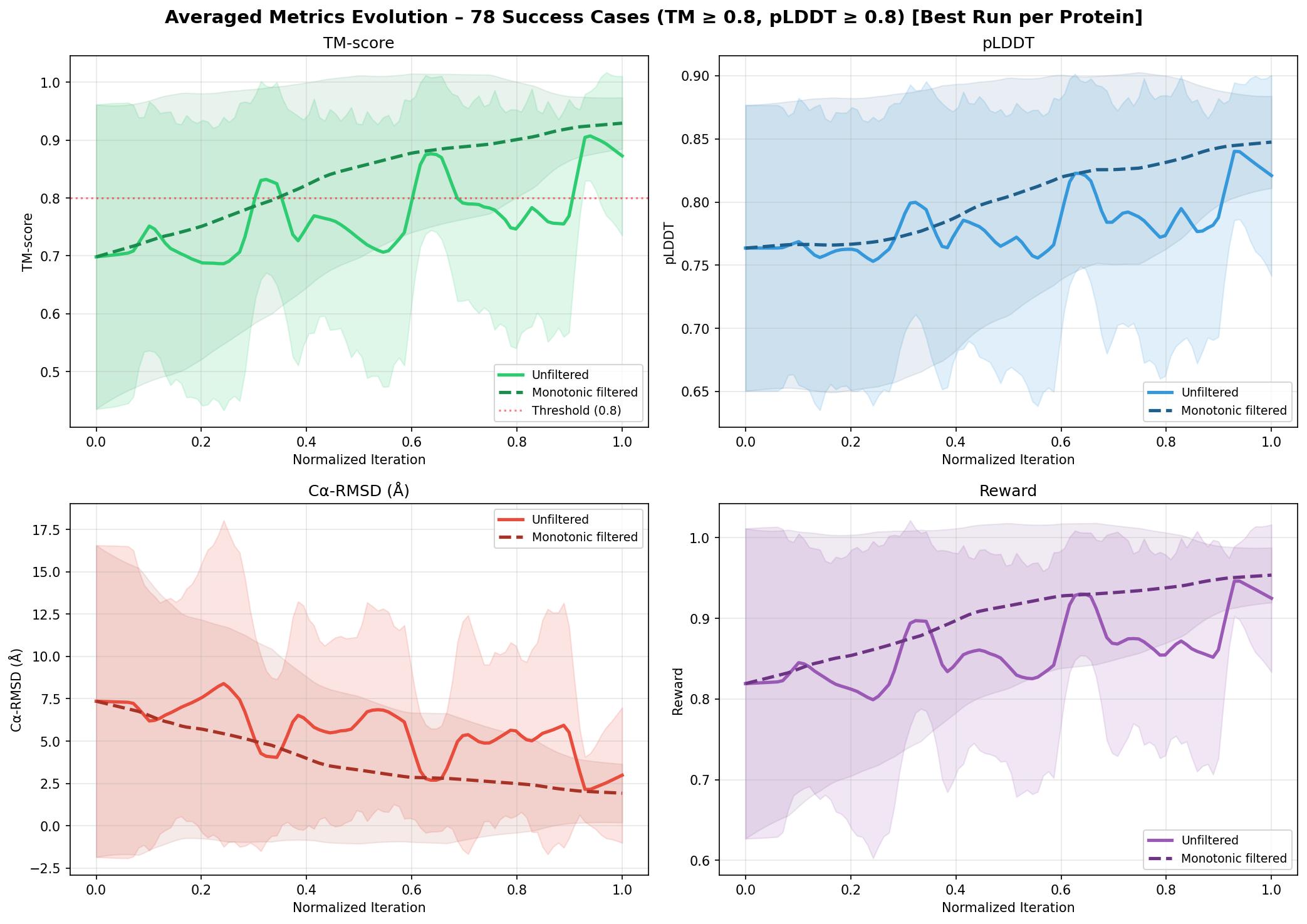}
    \caption{Averaged metric and reward evolution across 78 successful cases (TM-score $\geq$ 0.8, pLDDT $\geq$ 0.8) over normalized optimization iterations. Solid lines show unfiltered per-iteration values and dashed lines show 
monotonically filtered trajectories. Shaded regions show the standard deviation across proteins. These are RosettaSearch results with Priority Search algorithm and multi-objective optimization with all metrics.}
\label{fig:avg:metrics}
\end{figure*}

\subsection{Fidelity metrics improve over search steps}
Figure~\ref{fig:avg:metrics} shows the averaged evolution of all three fidelity metrics and the composite reward across 78 successful optimization trajectories. Rejection sampling of the trajectories produces monotonically filtered curves (dashed) that show consistent 
improvement across all metrics over the course of optimization, confirming that RosettaSearch performs meaningful search rather than benefiting from sampling luck. The unfiltered curves (solid) exhibit expected 
non-monotonic fluctuations at individual steps — 
reflecting the exploratory nature of the priority search -- while the overall trend remains consistently positive.

\subsection{Multi-modal extension with vision-language models}

As shown in Table~\ref{tab:vlm:results}, the VLM variant 
achieves comparable performance to the text-only (i.e. LLM) multi-objective variant when initialized from LigandMPNN designs (setting (b)). It achieves a success rate of 16.6\% under RF3 evaluation and 7.1\% under Chai-1 evaluation, compared to 17.4\% and 6.7\% respectively for the text-only 
multi-objective variant. However, it performs at par or marginally worse when initialized with native monomer sequences (setting (a)). 
Qualitatively, we observe that the VLM produces richer 
chain-of-thought reasoning, incorporating spatial and 
structural context from the images that is absent in 
text-only reasoning traces.

\begin{table*}[t]
\centering
\caption{Design success rates obtained by RosettaSearch with a vision-language model (VLM) with visual structural 
feedback, compared to the corresponding LLM performance while optimizing the two following objectives simultaneously: single sequence pLDDT and TM-score. Both approaches used the Sequential Revision algorithm as our VLM extension is currently limited to an iterative search.}
\begin{small}
\label{tab:vlm:results}
\renewcommand{\arraystretch}{1.1}
\begin{tabular}{p{2.2cm}p{1.6cm}p{5.5cm}p{1.2cm}p{1.2cm}}
\toprule
& & & \textbf{Success Rate (RF3)} & \textbf{Success Rate (Chai-1)} \\
\midrule
\textbf{Initialization} & \textbf{Algorithm} & \textbf{Approach} & & \\
\midrule
(a) Native & & unoptimized native protein sequences & 7.0\% & 2.4\% \\
monomer & Sequential- & LLM multi-obj. (ss-pLDDT, TM-score)   & \textbf{8.9}\%  & \textbf{4.7}\% \\
sequences & Revision & VLM multi-obj. (ss-pLDDT, TM-score)    & 8.3\%  & 4.6\% \\
\midrule
& & unoptimized LigandMPNN designs & 7.9\% & 3.1\% \\
(b) LigandMPNN & Sequential- & LLM multi-obj. (ss-pLDDT, TM-score)   & \textbf{17.4}\% & 6.7\% \\
designs & Revision & VLM multi-obj. (ss-pLDDT, TM-score)    & 16.6\% & \textbf{7.1}\% \\
\bottomrule
\end{tabular}
\end{small}
\end{table*}

\subsection{RosettaSearch improves BindCraft binders}
Next, we investigate whether RosettaSearch can improve fidelity of candidates generated by state-of-the-art approaches like BindCraft that involve expensive finetuning of large-scale foundation models like AlphaFold2. As seen in Table \ref{table:bindcraft}, we find that starting with the high-quality binder protein sequences generated by BindCraft, RosettaSearch is able to improve fidelity metrics in all 48 binders, by reasonable margins, while retaining the binding pockets involving the target enzymes.

\begin{table}[h]
\centering
\renewcommand{\arraystretch}{1.1}
\setlength{\tabcolsep}{5pt}
\caption{Improving BindCraft designed optimal binders} 
\label{table:bindcraft}
\begin{small}
\begin{tabular}{lcc}
\toprule
metric & BindCraft & RosettaSearch \\
& average & average \\
\midrule
ss-pLDDT $\uparrow$ & 0.85 & 0.90 \\ 
TM-score $\uparrow$ & 0.93 & 0.95 \\ 
C$\alpha$-RMSD $\downarrow$ & $1.37 \AA$ & $0.99 \AA$ \\ \hline
\end{tabular}
\end{small}
\vspace{-4mm}
\end{table}

\subsection{Human experts' thoughts on the LLM reasoning}
We examined the reasoning trace produced by the large language model for every proposal that it generates. We had 3 domain experts with varying degrees of experience and background in biochemistry, structural biology, biophysics, machine learning, directed evolution and protein engineering who were shown details of each optimization step and the reasoning as shown in Figure \ref{fig:expertanno}. 
We observed that depending on the feedback, the LLM either proposes targeted, minimal edits focused on low-confidence regions or generates an entirely new sequence that satisfies global design constraints such as length, residue diversity, and lack of internal repeats. 

Qualitatively\chingan{if quantitative, we need to give some numbers}, we evaluate results in a setting where the LLM is provided the native sequence as context.
The experts found the reasoning to have a more fixed pattern. The LLM focused on N-terminal regions that had low TM-scores. However, N- and C-terminal regions of predicted structures commonly exhibit poor alignment with the reference not because the sequence is infeasible, but because these regions are typically flexible and loop-rich, where close structural agreement is not expected.

Further, human experts agree that the sequence modifications suggested makes sense in isolation (e.g., replace buried charged residues with nonpolar ones) but that are not logical in the context (e.g., ``the residue that was changed from charged to nonpolar was not actually buried''). 

\meghana{Add results on vision settings}

\subsection{LLM experts' opinion of reasoning}
We used Gemini-3 as an independent evaluator of LLM reasoning quality. Specifically, three Gemini-3 experts were shown the reasoning output generated by o4-mini \chingan{we've not mentioned o4-mini is used} for each proposed protein sequence modification and asked to assess whether the reasoning was logically sound and consistent with the described sequence edits. Across 50 evaluated reasoning cases, the group of Gemini-3 experts were unanimous in their opinion where they marked 28 instances (56\%) as valid and 22 instances (44\%) as invalid.
%








\subsection{Evaluation on the Dayhoff dataset}
We show that our approach is not overfitting to the vagaries of the dataset or the sequence design approach involved, by evaluating it on another dataset which involves a subset of the Dayhoff atlas \cite{yang2025dayhoff}. 
The subset of Dayhoff that we use consists of 275 randomly sampled unconditionally generated RFDiffusion \textit{de novo} backbones, for which ProteinMPNN was used to generate sequences. Note that this dataset involves sequences that already have very good fidelity metrics (and hence success rates) as compared to the PDB dataset, which was curated based on low values on structural fidelity metrics. As before, success-rate-1 is calculated as pLDDT $\geq 0.8$, TM-score $\geq 0.8$ and success-rate-2 is calculated as pLDDT $\geq 0.8$, TM-score $\geq 0.8$, C$\alpha$-RMSD $\leq 1.5 \AA$. We find that RosettaSearch improves success-rate-1 from 72.4\% to 89.5\% and success-rate-2 from 45.5\% to 67.3\%.

An important distinction from the PDB monomer 
evaluations is that no reference sequence is provided 
as context to the LLM optimizer in any of the experiments in this setting -- 
the LLM must propose sequence edits based solely 
on structural feedback from RosettaFold3, without 
any reference sequence information. This makes the 
Dayhoff evaluation a stricter test of the LLM's 
ability to interpret and act on structural feedback 
alone, demonstrating that RosettaSearch does not 
rely on reference sequence context to achieve 
meaningful improvements in structural fidelity.

\subsection{Results from ablation of the contextual signals} 
\label{appendix:ablation}
Table \ref{tab:ablation:study} presents an ablation study that isolates the contribution of different forms of context and feedback in our optimization framework, using LigandMPNN-designed sequences as the starting point. 
Introducing reward-only sequential revision yields a minor improvement to 9.3\%, indicating that scalar structural rewards alone provide limited guidance. Augmenting with detailed textual feedback gives a modest improvement to performance to 11.4\%, highlighting the importance of rich, interpretable feedback. Providing the native reference sequence in context leads to a large jump in success rate to 17.2\%. Surprisingly, there is no further gain when protein functional annotations are included (17.4\%). Finally, incorporating retrieval-augmented generation (RAG) from a larger dataset yields slightly lower performance (16.1\%). Given these observations, we provide at the most, the native protein sequence as context, which also has the benefit of limiting the number of input tokens to the LLM.

\begin{table}[h]
\centering
\renewcommand{\arraystretch}{1.1}
\setlength{\tabcolsep}{5pt}
\caption{Evaluation on Dayhoff backbones where ProteinMPNN was used to generate candidate sequences. RosettaSearch results are shown with multi-objective optimization. \meghana{TODO: report statistical significance}} 
\label{table:dayhoff}
\begin{tabular}{lcc}
\toprule
Metric & ProteinMPNN & RosettaSearch \\
& average & average \\
\midrule
ss-pLDDT  $\uparrow$ & 0.81 & 0.85 \\
TM-score $\uparrow$ & 0.90 & 0.96 \\ 
C$\alpha$-RMSD $\downarrow$  & $3.32 \AA$ & $1.78 \AA$ \\ \hline 
success rate-1 & 72.4\% & 89.5\% \\ 
 & (199/275) & (246/275) \\ 
success rate-2 & 45.5\% & 67.3\% \\ 
 & (125/275) & (185/275) \\ 
 \bottomrule
\end{tabular}
\end{table}

\begin{table}[t]
\centering
\caption{Results of an ablation study on context, showing the baseline success rate and the successive improvements. The Sequential Revision baseline was used for this evaluation.}
\label{tab:ablation:study}
\renewcommand{\arraystretch}{1.3}
\setlength{\tabcolsep}{6pt}
\begin{small}
\begin{tabular}{l|c} \hline
LigandMPNN designed sequences  & 8.5\% \\ \hline
$\indent$ + reward-only sequential revision & 9.3\% \\ \hline
$\indent$ $\indent$ + detailed feedback text & 11.4\% \\ \hline
$\indent$ $\indent$  $\indent$ + native sequence in context & 17.2\% \\ \hline
$\indent$ $\indent$ $\indent$ $\indent$ + protein function in context & 17.4\% \\ \hline
$\indent$ $\indent$ $\indent$ $\indent$ + RAG from bigger dataset & 16.1\% \\ \hline
\end{tabular}
\end{small}
\vspace{-5mm}
\end{table}

\section{Further results}
Results showing the improvements in fidelity as a function of sequence length are shown in Figure \ref{fig:protein:length}, where \textit{short} refers to sequence lengths of less than 100 amino acids, a length of 100 to 350 is considered \textit{medium} length and greater than 350 amino acids is considered a \textit{long} protein. Our PDB dataset with $\approx$400 structures has 249 medium, 129 long and 36 short monomers. We find that the improvements obtained by RosettaSearch in pLDDT and TM-score are the highest for medium length proteins, whereas the improvements in C$\alpha$-RMSD are the highest for short proteins.

For additional results, please refer to the Appendix.

\section{Conclusion}
\chingan{This is crucial. After reading the experiment section, I think people will get lost. We need to have a summary of findings. Importantly, we need to point out the significance (we should repreat this multiple times, in abstract, intro, exp sec, and here. In addition, we should point out limitation. }

\chingan{Btw, we should list hyerparameter choices and compute spent for these experiments in the appendix.}

Our experimental results demonstrate that reasoning-guided priority search enables reliable and scalable improvements in protein sequence fidelity. By combining structured feedback from RosettaFold3, multi-objective optimization, 
and priority-based parallel candidate exploration, RosettaSearch yields consistent structural improvements across datasets and initialization regimes, outperforming both sequential revision strategies and repeated sampling baselines. Importantly, relative gains are preserved under evaluation with another independent structure predictor (Chai-1) and generalize across two distinct 
LLM families (o4-mini and Gemini-3), supporting the robustness and generality of the framework. Taken together, these results suggest that LLM-driven reasoning and search can serve as a powerful complement to existing protein 
design pipelines, particularly in regimes where single-shot inverse folding struggles to recover highly designable sequences.

A current limitation of our approach is its focus on optimizing a single sequence per target backbone rather than producing a diverse set of high-fidelity candidates, which is a desirable property for practical protein design workflows. Addressing this through diversity-aware reward 
formulations and explicit diversity objectives in the search algorithm is an important direction for future work.



\paragraph{Code and Data Availability}
All code, data, generated sequences will be made available on github.

\section{Acknowledgements}
We thank Bruce Wittman, Eric Horvitz, Ethan Jackson, Chase Harms, and David Baker for support, discussions, and providing feedback on this work.


\bibliography{references}
\bibliographystyle{icml2026}

\newpage
\appendix
\onecolumn

\section{Appendix}

\subsection{Calculation of $C\alpha$-RMSD}
\label{appendix:calpha}
To measure structural fidelity $C\alpha$-RMSD $(\mathbf{x},\mathbf{x}^*)$ of $\hat{\mathbf{x}}$ to the provided reference native structure or reference backbone $\mathbf{x}^*$, where $L = L^*$ \chingan{what's the difference between reference native structure or reference backbone, can they be use interchangeably here - DONE?}, we superimpose the predicted structure of the generated proposal over the reference structure using an algorithm \cite{zhang2005tm} that aligns remotely homologous protein structures (pairs of proteins whose sequences have sequence identity less than 30\%). We use this approach since the generated sequence could have an arbitrarily low sequence similarity to the native sequence.
\chingan{Not sure if I fully understand the final two sentences. Why is superimposition needed? Is it needed when $L=L^*$? What are the implications of using it? - domain knowledge}

\subsection{Multi-objective optimization details}
\label{appendix:multiobj}
First, we evaluated a round-robin scheme in which optimization alternates between objectives across iterations, allowing the model to focus on a single criterion at a time while implicitly encouraging coverage of the full objective set. 
Candidate evaluations are ranked using a single objective at a time, with the active objective cycling across searches to encourage balanced coverage of the objective space.

Second, we experimented with scheduled reward formulations in which the relative emphasis on different objectives is varied over the course of optimization, including curricula that prioritize coarse global structure in early iterations and increasingly emphasize local refinement signals in later iterations. With priority search which works in parallel, the priority assigned to different objectives varies across evaluations based on a predefined curriculum. 


\begin{figure*}[b]
    \centering
    \label{fig:plddt:metrics}
    \caption{Results showing improvements in the various metrics}
    \begin{subfigure}{0.38\linewidth}
        \centering
        \includegraphics[width=0.8\linewidth]{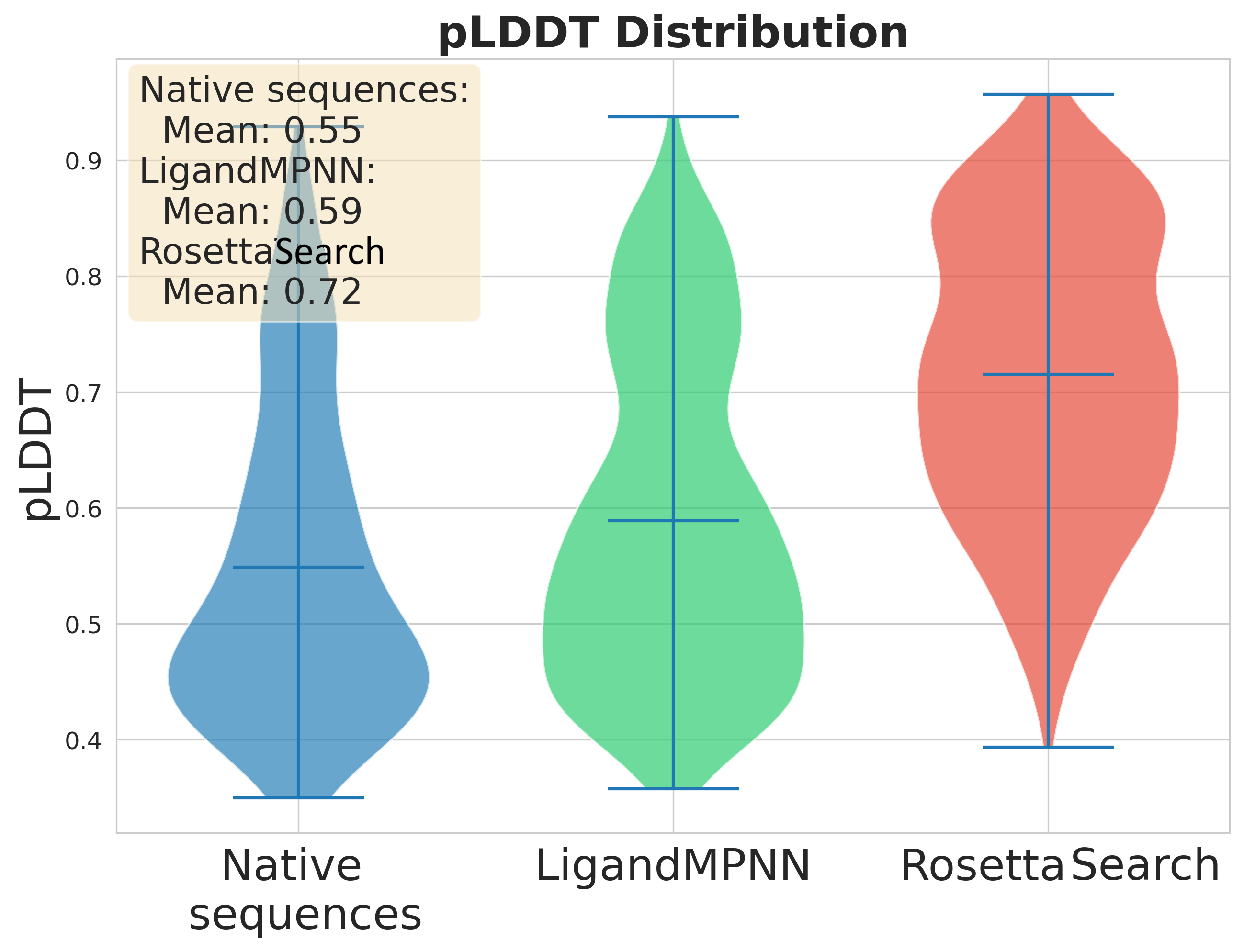}
        \caption{Distribution of pLDDT values of the baseline LigandMPNN designs that were used as starter sequences for RosettaSearch to do single objective optimization of pLDDT and the obtained improved pLDDT values.}
    \label{fig:plddt:improvement}
    \end{subfigure}
    \hspace{0.2cm}
    \begin{subfigure}{0.58\linewidth}
        \centering
        \includegraphics[width=\linewidth, height=4cm]{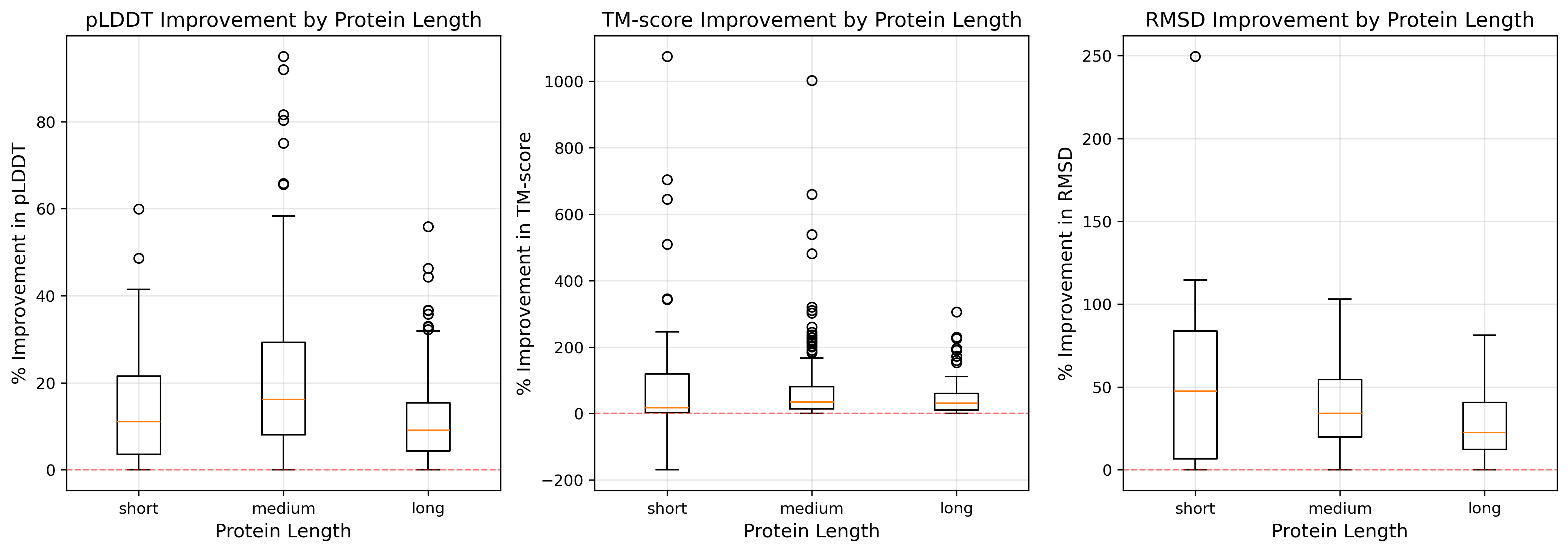}
        \caption{Stratifying the \% improvement in fidelity metrics by the length of the protein sequence. (\textit{Left}) pLDDT improvement, (\textit{Center}) TM-score improvement, (\textit{Right}) $C\alpha$-RMSD improvement. Results are shown for Search with multi-objective optimization.}
        \label{fig:protein:length}
    \end{subfigure}
\end{figure*}

\begin{figure*}
    \centering
    \includegraphics[width=\linewidth]{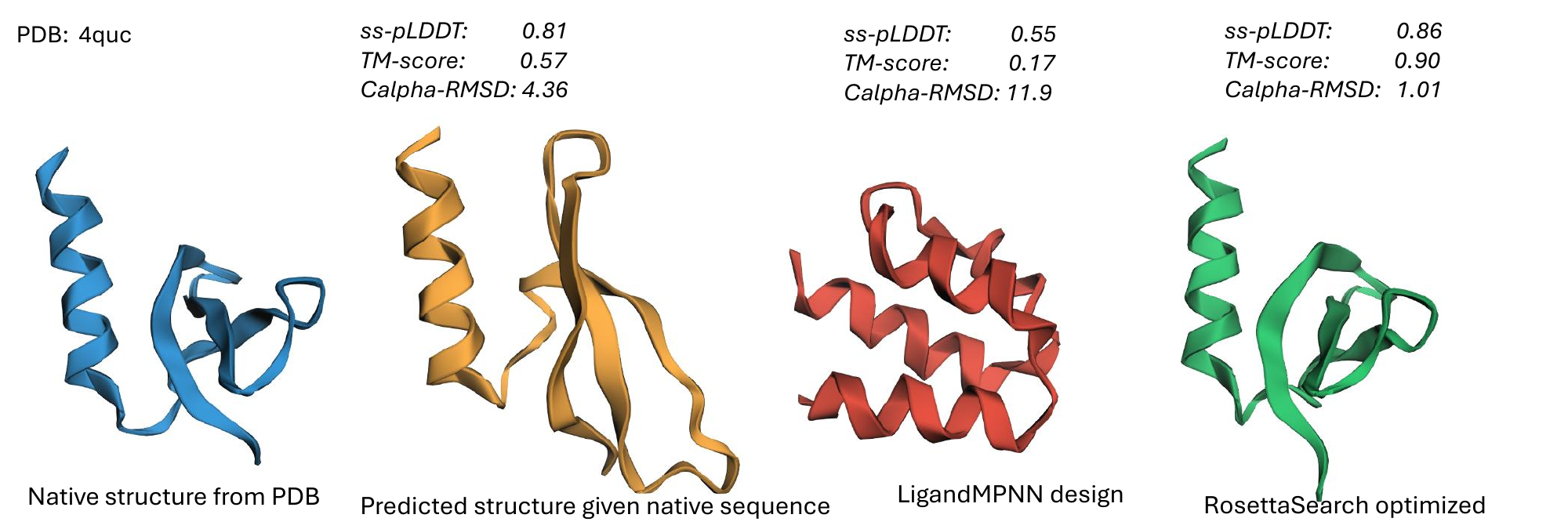}
    
    \vspace{1em}
    
    \includegraphics[width=\linewidth]{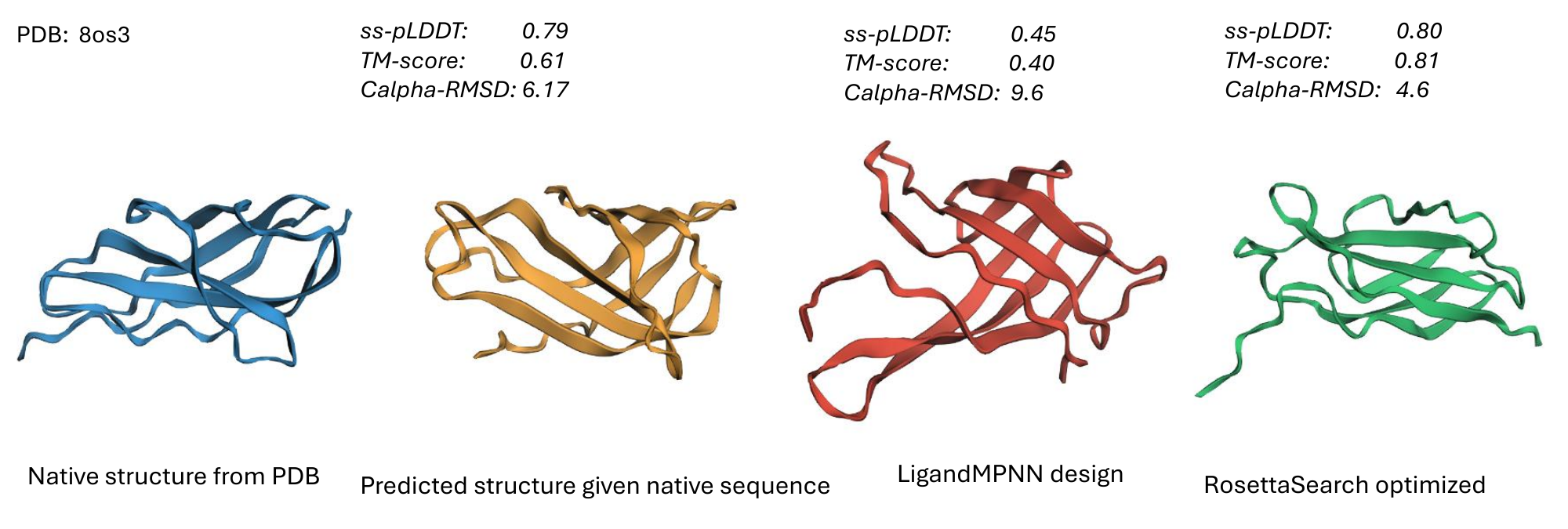}    

    \vspace{0.5em}    
    
    \includegraphics[width=\linewidth]{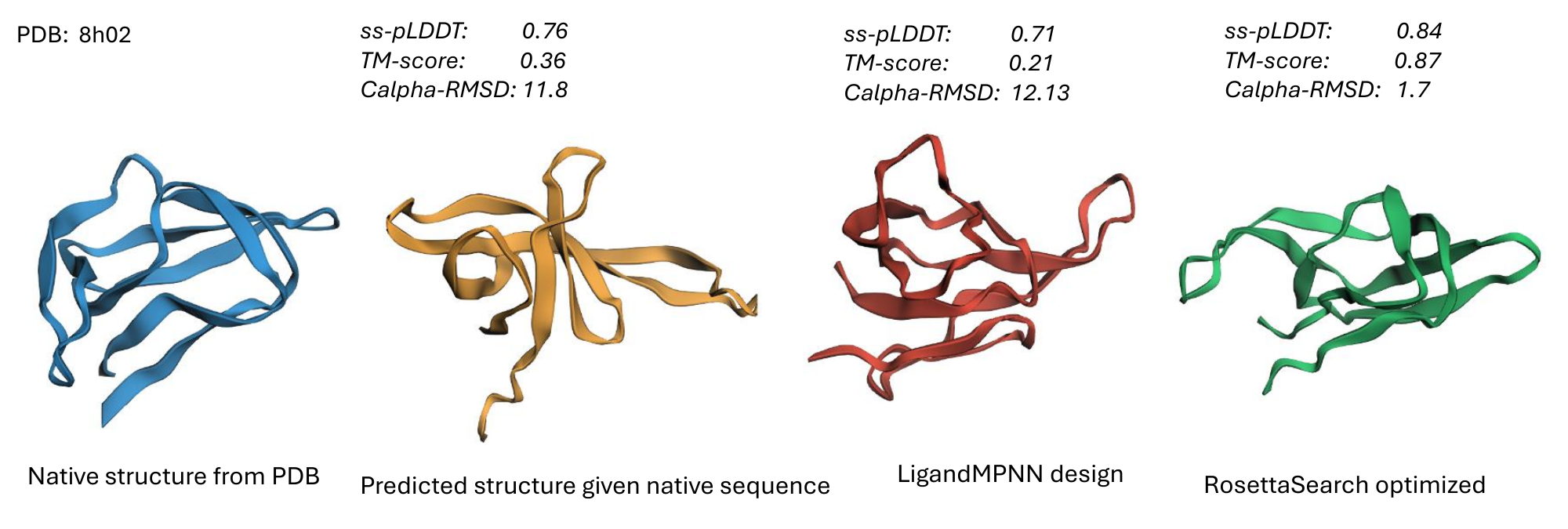}        
    
    \caption{More examples of structures from the PDB dataset that were optimized using RosettaSearch successfully. See Table \ref{tab:seqs} for the amino-acid sequences corresponding to these structures.}
    \label{fig:structure:res}
\end{figure*}

\begin{table}[h]
\centering
\setlength{\tabcolsep}{3pt} 
\renewcommand{\arraystretch}{1.05} 
\footnotesize
\begin{tabularx}{\columnwidth}{p{2.4cm}p{4.7cm}p{4.7cm}p{1.4cm}}
\toprule
\textbf{Dayhoff backbone id} & \textbf{ProteinMPNN design from Dayhoff atlas} & \textbf{RosettaSearch optimized} & \textbf{Sequence identity (blastp)} \\
\midrule
    \texttt{20240891510\_ CvNknz\_ 00140AA\_2} & \seqsplit{AAEKEKEELKKAREKMEKEIKELVEKLKGYIIISILTAVLREVSGKLVVEIKVVAGSESIKLEELEAKMKEYIAKKKAEFAALGLKVEEEIKRSEKEVVAKLTAPLDKLPPESDKLKRTLEIINLETGEVEVRKEEITLE} & 
\seqsplit{MKAKQKLALKKAREKMALQIKELVAKLKGYIIISILTAVLREVTAKLVVEIKVVAGSASIRLEELEAKMKEYIAKKKAEFAALGLKVEEEIKPSPEEVEVKLTAPLDKLPESDTLKRTLEIINLETGEVEVRKEEITLAA} & 88.2\%
\\ 
\texttt{20240891627\_ ZYl8lJ\_ 00304AA\_2} 
&  
\seqsplit{ATLELTGPLTPEFLAKARELLKPLAAAEEALTALDEVVALGNATVTIYKQQGWTTTSTLTLTIERRPNEEKVTVTLTVTATAADGTTRTTTITATVTYVELPGGPPVLKRTLEEKASDGSSFTLSISTTGAMELSLQVPGKEPVSLTLDIIKGGELERKFYVDPSSGNRIIVIKATKDHIVVIIRIRKLTPESKELLKEILEEVFKEAKKEGAKVHVVILADTLENAKIALDIVLEQALKEGVTLSISLGVAVKAEEADEVYEELVKYVEEKLEKIREEGKVKVERVEVRLPGKTAVFEVEEEI}
& 
\seqsplit{MTLENTAALTAEFLAKARELLKPLAAAEEALTALDEVVALGLATVTILALQGWTTTSTLTLTIERKANEEKVTVTLTLTALAANGTTRTTTITATVTYVELPGGPPVLKRTLEEKASDGSSFTLSISTTGAMELSLQVPGKEPVSLTLDIITGGELERKFYVDPSSGNRIIVIKATKDHIVVIIRIRKLTPESKELLKEILEEVFKEAKKEGAKVHVVILADTLENAKIALDIVLEQALKEGVTLSISLGVAVKAEEADEVYEELVKYVEEKLEKIREEGKVKVERVEVRLPGKTAVFEVEEEI} & 95.4\% \\
\texttt{20240891538\_ B8FFVA\_ 00355AA\_0} 
&
\seqsplit{TAQLQADLDRVVAALAANPRIQAIRAILAQRPEVEPEAVSNERALELVMEVLGLRDRAEARALMVPAARLFGVPVPPLSPEEQELLAAALARPVLTVALVFRVRYSRFEVLEEEEREYEDPERPGLRVRERRRVVRVDGYLYRFVHIEVRDEATGKVVRAGLLSVQVGPDHYVISVDFEAEPGASEEAVDLVVDRVNEELREIAKEIKKATTLSLSIRDTAHAPKVIQAFVAAANEIAPYVKSLKVSIHVGKHPGPITLTLTLTPGVDALTVIAENCAELNVNVTAGVKELRFTAYSTKITVTLAPDAKIEKVEFVETTPETVINTIAEVVLECLATEEDALGVVVLREEVGAVI}
&
\seqsplit{TAQLQADLDAVVAALAANPEIQAIRAILAQRPEVLPEAVSNERALELVMIVLGLRDRAEARALMVPAARLFGVPVPPLSPEEQELLAAALARPVLTVALVFRVRYSRFEVLEEEEREYEDPERPGLRVRERRRVVRVDGYLYRFVHIEVRDEATGKVVRAGLLSVQVGPDHYVISVDFEAEPGASEEAVDLVVDRVNEELREIAKEIKKATTLSLSIRDTAHAPKVIQAFVAAANEIAPYVKSLKVSIHVGKHPGPITLTLTLTPGVDALTVIAENCAELNVNVTAGVKELRFTAYSTKITVTLAPDAKIEKVEFVETTPETVINTIAEVVLECLATEEDALGVVVLREEVGAVI}
& 98.9\% \\
\bottomrule
\end{tabularx}
\caption{Examples of ProteinMPNN sequences from the Dayhoff atlas that were redesigned using RosettaSearch, corresponding to the structures from Figure \ref{fig:structure:res:dayhoff}.}
\label{tab:seqs:dayhoff}
\end{table}

\begin{table}[h]
\centering
\setlength{\tabcolsep}{3pt} 
\renewcommand{\arraystretch}{1.05} 
\footnotesize
\begin{tabularx}{\columnwidth}{@{} l >{\RaggedRight\arraybackslash}X >{\RaggedRight\arraybackslash}X >{\RaggedRight\arraybackslash}X @{}}
\toprule
\textbf{PDB id} & \textbf{Native sequence} & \textbf{LigandMPNN design} & \textbf{RosettaSearch optimized} \\
\midrule
    \texttt{7z05} & 
    \seqsplit{KESLTHASVLSAYTASYKTTAIKSIADNAVVLDIGYGKGNDGPRYAVRPLTVTGIDTAARMLAIADQNKPENVTLVKQGFFTHITKTSNTYTHVIAFNSLHYPLASSHPDTLVQRLPTCPANILIPCHHLLEGIQTPTYSVVKDEDMWCVKVTKNEFIESSYNYDVFVKALESKYHVTIGSLLDCVEKPSTRSITPTLWTAMRNFVNNDQEMQRILSGYITFNLTPLPPKVEIINDWLD} & 
    \seqsplit{EDGLTHADVLFAFRGSAQTRAVDAIPEGSTVLDIGVGTGRLFPLYARRPLTVTGIDVDAAALAKAEAHRPANVTLVHQDLFTYLKETDVEYTYVLAFDALHYPLSRSDPRELVDLLPRAPAFILIPDFRLLEGVETPTFSVRPDGDRVLVRIGGRELTYDYYDLDAFEAALAERYDVRRGTLLDETTKPADPSITDELWERMRALVANDEDLQRILGGFRTYDLTPKPPPAAAAAAAAA} &  
\seqsplit{MESLTHADVLFAYTASYKTTAIDAIPEGSTVLDIGVGTGRLFPLYARRPLTVTGIDVDAAALAKAEAHRPANVTLVHQDLFTYLKETDVEYTYVLAFDALHYPLSRSDPRELVDLLPRAPAFILI PDFRLLEGVETPTFSVRPDGDRVLVRIGGRELTYDYYDLDAFEAALAERYDVRRGTLLDETTKPADPSITDELWERMRALVANDEDLQRILGGFRTYDLTPKPPPAEIINDWLD}
    \\ \hline
    \texttt{4quc} & \seqsplit{EYVVEKILGKRFVNGRPQVLVKWSGFPNENNTWEPLENVGNCMKLVSDFESEVFRL} & \seqsplit{SADPDKILGRLLRNGFPYYHIHWLHMTQENITWRSAAHLGNIGFQWILFRYALQRK} & \seqsplit{EYVVEKILGKRFVNGRPYYHIKWL GMTQENITWRSAANVVGCMKLVSDFESELQRK}  \\ \hline
    \texttt{8os3} & \seqsplit{GSSSPPSPPLDLHVTDAGRKHIAIAWKPPEKNGGSPIIGYHVEMCPVGTEKWMRVNSRPIKDLKFKVEEGVVPDKEYVLRVRAVNAIGVSEPSEISENVVAKD} & \seqsplit{GEVAAPTEPTSLKVTAASKDYIVIEWQSPESDGGTPKRGYSVQMVRVFTENWVEVNSALIKSNSYQVTAGVTPNEVFKLRVTAFNDAGSSEPSQESENVKKAT} & \seqsplit{GSAAAPTEPTSLKVTAASKDYIVIEWQSPEKNGGSPILGFSVQMVRVFTENWVEVNSRPIKSNSYQVTAGVTPNEVFKLRVTALDQAGSAEPSEESENVVKAT} \\ \hline
    \texttt{8h02} & \seqsplit{TARLLRAPVAGTIKLGKKARTRPYRTRHGEEALLAEANFDLVLEGKGRKETFAILQGSTIFVQDGDKVAAEAILAEVPV} & \seqsplit{DYEVIFAHEPGFKILGDEAVLERYVTKDGNDVLRAKKNFTLIIVGKKKKHESNGPEGSGVLVRDGDTVDKGEPLAIVPL} & \seqsplit{TARLLRAPVAGTIILGDEAVLERYVTKDGNDVLRAKKNFTLIIVGKGRKETFNGPEGSGVLVRDGDKVAAGEPLAIVPL} \\ \hline
\bottomrule
\end{tabularx}
\caption{LigandMPNN sequences that were redesigned using RosettaSearch. The original native PDB monomer sequence that was being redesigned is also shown. The sequences shown here correspond to the structures shown in Figure \ref{fig:struc:res} and Figure \ref{fig:structure:res}.}
\label{tab:seqs}
\end{table}

\begin{figure*}
    \centering
    \includegraphics[width=\linewidth]{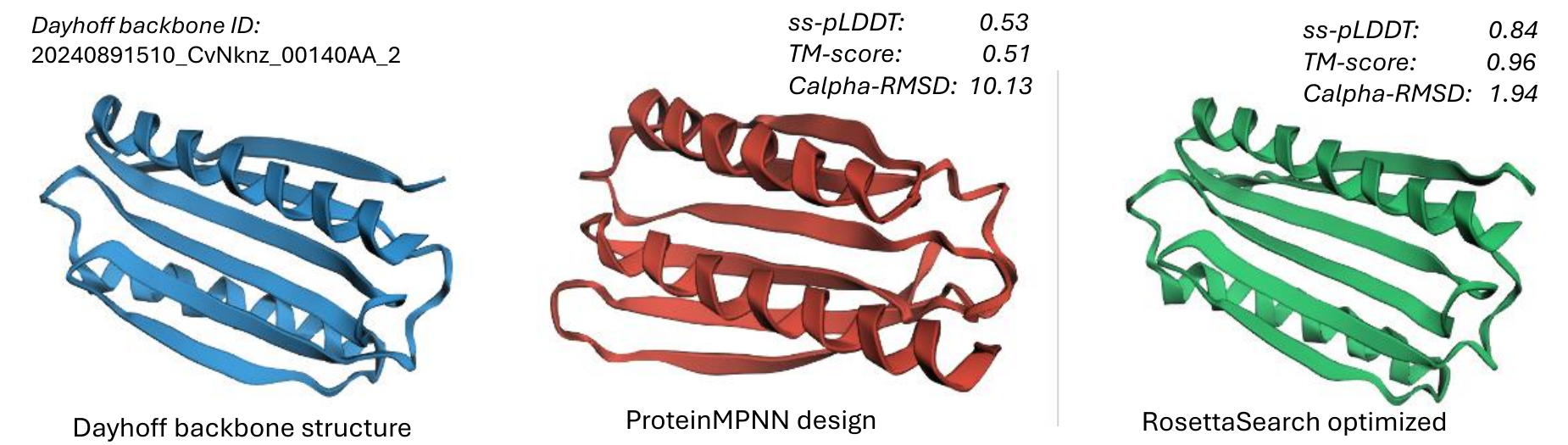}
    
    \vspace{2em}
    
    \includegraphics[width=\linewidth]{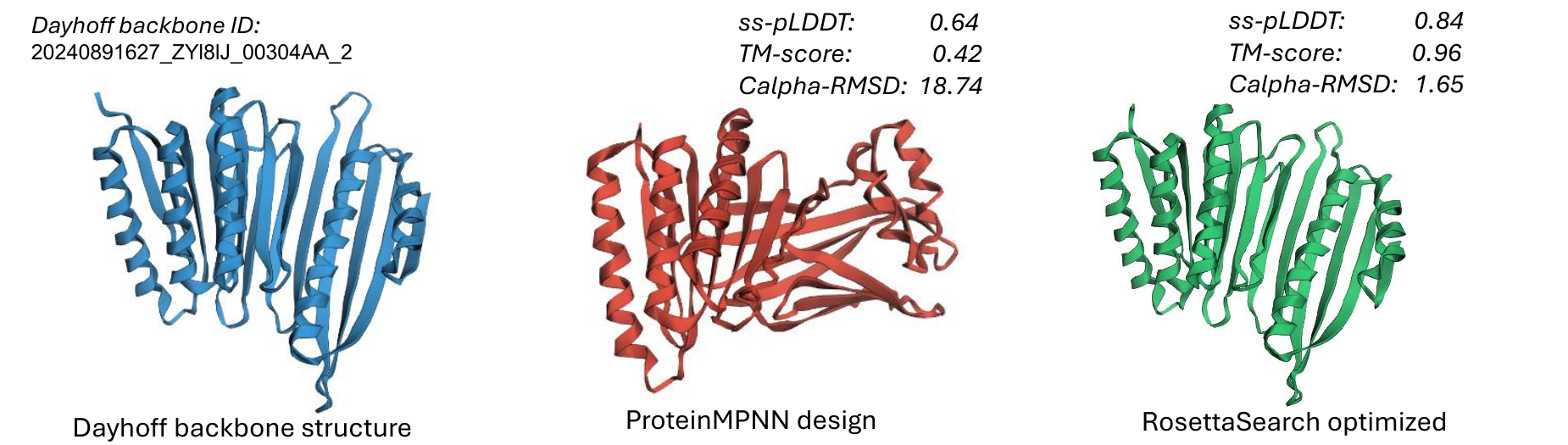}    

    \vspace{2em}    
    
    \includegraphics[width=\linewidth]{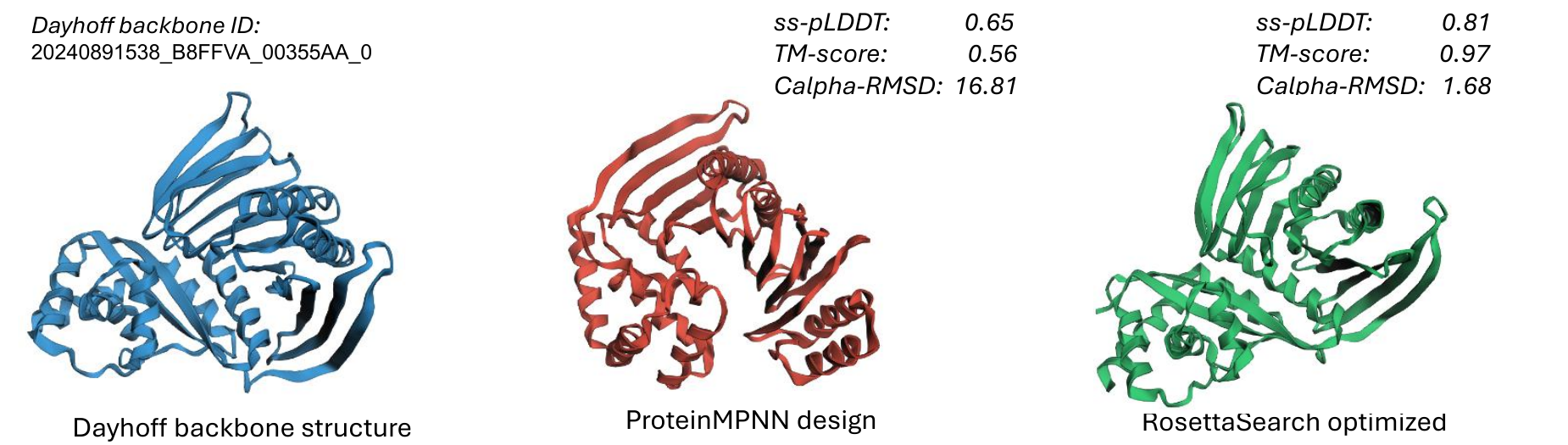}        
    
    \caption{Examples of structures from the Dayhoff Atlas that were optimized using RosettaSearch successfully. See Table \ref{tab:seqs:dayhoff} for the amino-acid sequences corresponding to these structures.}
    \label{fig:structure:res:dayhoff}
\end{figure*}

\begin{algorithm}[t]
\caption{Search}
\label{alg:priority-search}
\begin{algorithmic}[1]
\REQUIRE 
$\phi_0$: starting protein sequence \\ 
$\mathcal{O}$: LLM Optimizer\\
$G$: Guide that provides feedback
(which provides feedback about pLDDT, TM-score, $C\alpha$-RMSD) \\
$K$: number of candidates considered for exploration\\
$N$: number of proposals generated per candidate\\
$S(\cdot)$: score function to prioritize candidates 

\STATE Initialize priority queue $\mathcal{M}$ which store candidates of parameters and their scores (i.e. $c = (\phi, \text{score})$.
\STATE Insert $K$ copies of $\phi_0$ into $\mathcal{M}$ with maximal priority

\FOR{iteration $t = 1, 2, \dots$}
    \STATE \textbf{(Exploit)} Select best candidate $\phi_t^*$ of
    $
        c^* \leftarrow \arg\max_{c \in \mathcal{M}} S(c)
    $

    \STATE \textbf{(Explore)} Select top-$K$ candidates from $\mathcal{M}$ as $\mathcal{C}$

    \FOR{each candidate $c \in \mathcal{C}$}
        \STATE Run $c$ using guide $G$ to extract score and feedback
        \STATE Collect rollouts $\mathcal{R}_c = \{(\phi, \text{score}, \text{feedback})\}$, where $\phi \in c$.
    \ENDFOR

    \STATE \textbf{(Propose)} Initialize proposal set $\mathcal{P} \leftarrow \emptyset$
    \FOR{each candidate $c \in \mathcal{C}$}
            \STATE $f \gets \bigcup \{ \text{feedback} | \text{feedback} \in r, r\in \mathcal{R}_c \} $
            \FOR{$p = 1$ to $N$}
                \STATE $\phi' \leftarrow \text{optimizer.step}(\phi, f)$
                \STATE Evalute $\phi'$ with guide $G$ to get $\text{score}'$
                \STATE Add candidate $c'=(\phi', \text{score}')$ to $\mathcal{P}$
        \ENDFOR
    \ENDFOR

    \STATE \textbf{(Memory Update)}
    \FOR{each candidate $c \in \mathcal{C} \cup \mathcal{P}$}
        \STATE Compute priority $S(c)$
        \STATE Insert $c$ into priority queue $\mathcal{M}$
    \ENDFOR
\ENDFOR

\STATE \textbf{return} best designs $\{\phi_t^*\}$
\end{algorithmic}
\end{algorithm}

\subsection{Where are the most updates made?}
\label{appendix:updates}
To understand whether there is a propensity in the substitutions made by the LLM, we analyze the positional information where the updates are made by dividing each protein sequence into four bins: N-terminal representing the beginning (0-25\%) of the sequence, two bins to represent the middle regions and C-terminal to represent the end of the protein sequence. We compare the final ``successful'' generated sequence (i.e. ones that satisfied the success criterion) to the starting sequence and calculate the number of updates that fall into each bin. As seen in Figure \ref{fig:pos:dist}, at 30\% the C-terminal sees the highest percentage of updates whereas the other regions have between 22 to 24\% of the updates.

\begin{figure}
    \centering
    \includegraphics[width=0.9\linewidth]{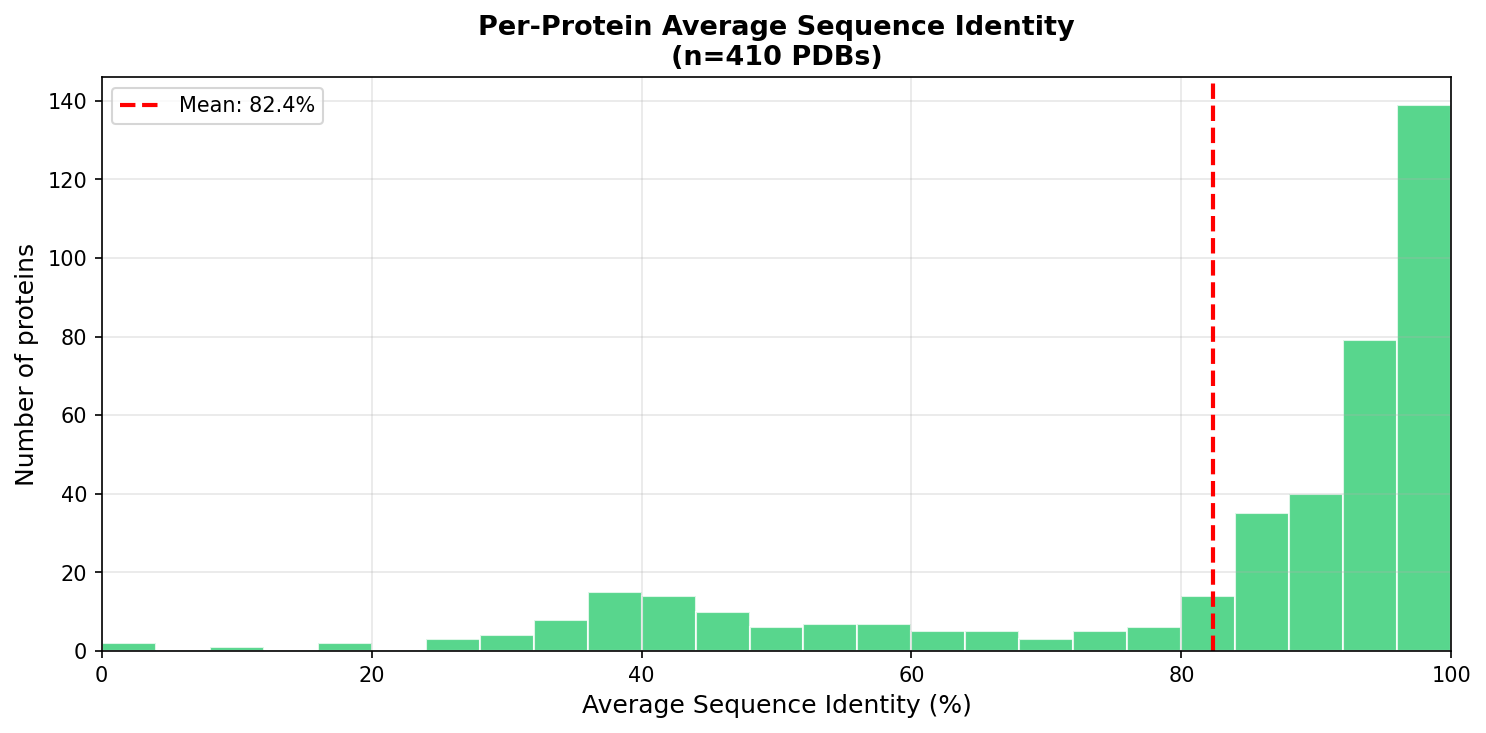}
    \caption{Distribution of the percentage sequence identity between the initial starting sequence and the generated protein sequences on the LigandMPNN dataset with 400 monomers.}
    \label{fig:seqid}
\end{figure}

\begin{figure}[t]
        \centering
        \includegraphics[width=0.7\linewidth]{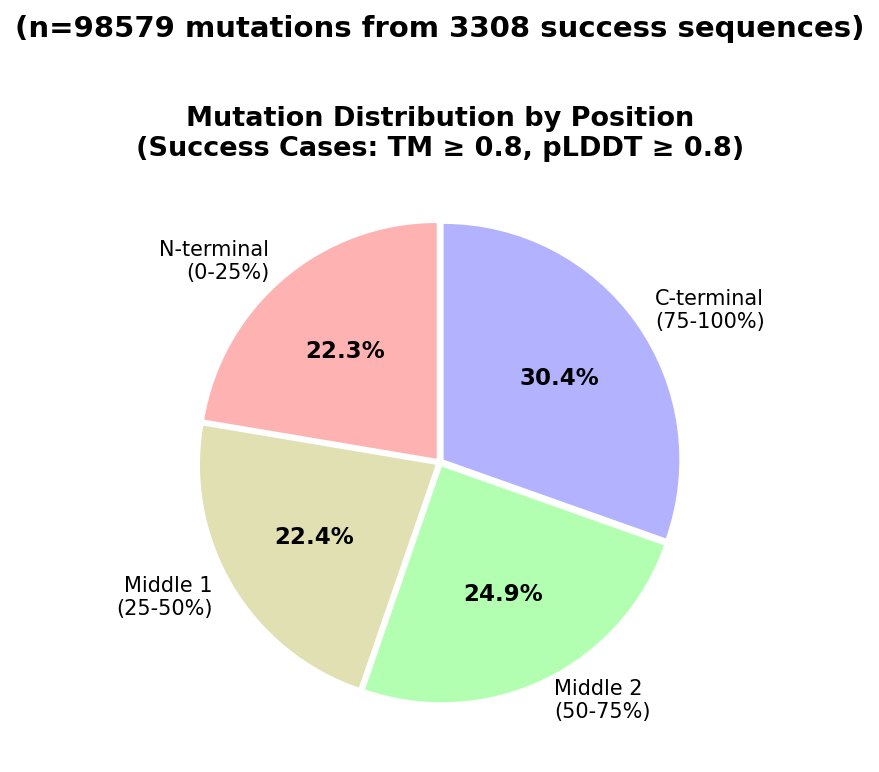}
        \caption{Distribution of positions that were updated in generated sequences as compared to the starting sequence. There were a total of 3308 generated sequences that satisfied the success criterion ((TM-score$\geq 0.8$) and (pLDDT$\geq 0.8$)) over the 400 proteins' roll-out trajectories explored with the LigandMPNN designs dataset.}
    \label{fig:pos:dist}
\end{figure}

\begin{figure*}
    \centering
    \includegraphics[width=\linewidth]{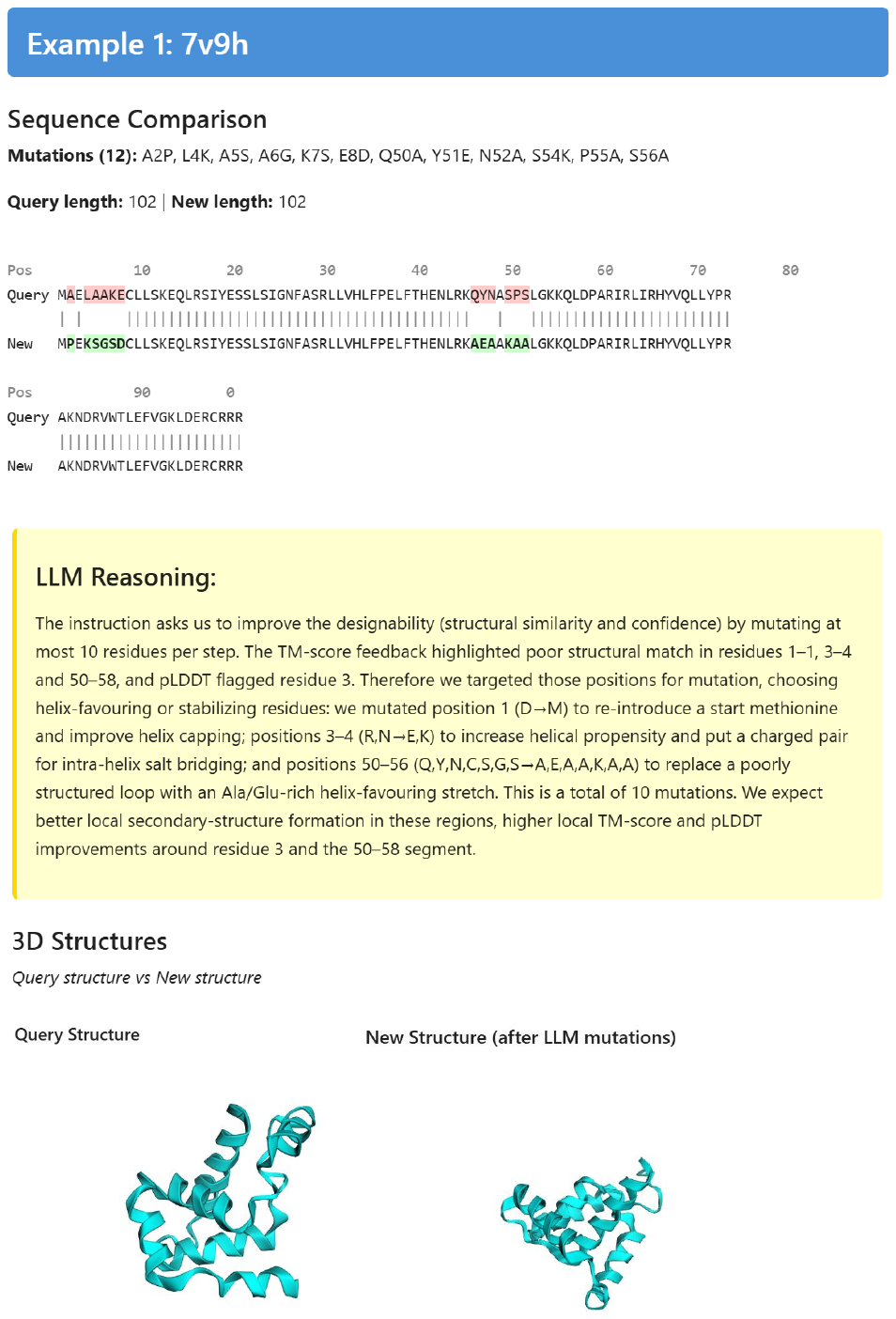}
    \caption{A screenshot of the expert annotation setup that also shows the LLM reasoning for one step of the protein sequence optimization. This is a randomly chosen example.}
    \label{fig:expertanno}
\end{figure*}

\begin{figure*}
\caption{Prompt input to the LLM to describe the task of protein sequence optimization for a given input sequence}
\label{fig:prompt}
\begin{tcolorbox}[
  title={Example Optimization Prompt},
  colback=gray!5,
  colframe=black,
  fontupper=\ttfamily\small,
  left=2mm,
  right=2mm,
  top=1mm,
  bottom=1mm
]
\textbf{\#Instruction}\\
We want to increase the structural fidelity of the proposed protein sequence, by changing a maximum of 20 residues in each step. The constraint is to match the fold of this native protein sequence: \\
\seqsplit{KSKLAKPIQDLIKMIFDVESMKKAMVEFEIDLQKMPLGKLSKRQIQSAYSILNEVQQAVSDGGSESQILDLSNRFYTLIPHDFGMKKPPLLSNLEYIQAKVQMLDNLLDIEVAYSLLRGGNEDGDKDPIDINYEKLRTDIKVVDKDSEEAKIIKQYVKNTHAATHNAYDLKVVEIFRIEREGESQRYKPFKQLHNRQLLWHGSRTTNFAGILSQGLRIAPPEAPVTGYMFGKGIYFADMVSKSANYCHTSQADPIGLILLGEVALGNMYELKNASHITKLPKGKHSVKGLGKTAPDPTATTTLDGVEVPLGNGISTGINDTCLLYNEYIVYDVAQVNLKYLLKLKFNYKTS}.  \\
Use only standard amino acids (ACDEFGHIKLMNPQRSTVWY), no gaps. Output only the raw sequence without any code.\\

\textbf{\#Variables}\\
(str) protein\_sequence=
\seqsplit{MSKLAPPIVELIKDIFDRERLEEELRRMQIDLERLPLDSLSEEMIDRALRILTEVSRVVKNGGSEEEIERLSEEFYALLPRDFGDEEPPLLDDPELIRQRVLDLDHLRRIRTTWRLLNGGLDSKEKHPIDILYEKLRTRIEVVDPDSEEAKLIRKYVENTWGPDDNDFDIEVEEIFRIEREGAAERFAPYLKKENRWLLFHGTRITELAGILRDGLTLPPPEAPIEGYMFGKGIEFTSVVSKAAKNMETSRENPIGYLLLAEVALGKMYKIKKPENIKELPEGYDSVHGIGKYAPDPKETTTLDGVKVPLGKLVETGETDTFFKYDRFVVYDEAQVLLRYLLKLKFNFYGD} \\

\textbf{\#Constraints}\\
(str) A protein sequence of 351 amino acid residues. Use only standard amino acids: ACDEFGHIKLMNPQRSTVWY. Do not include repeated patterns or gaps or motifs. Output ONLY the raw sequence without any code, or formatting. Example format: MKTAYIAKQRQISFVK...\\




\textbf{\#Feedback}
TM-score feedback: The TM-score is low, indicating poor structural similarity. Please make substantial improvements. The following residue regions have low structural similarity to the native structure (local average deviation >5.0 Å): 1-29, 32-41, 44-44, 46-53, 63-64, 72-74, 78-125, 140-157, 161-212, 217-267, 271-286, 290-308. Focus on improving these regions. pLDDT feedback: The following ranges of residues have low pLDDT (less than 35), indicating low confidence in the predicted structure: 111:111, 121:121, 137:137, 139:139, 180:180, 196:201, 204:206, 212:214, 216:222, 224:226,
 228:237, 239:239, 247:247, 251:252, 255:255, 257:257, 262:263, 265:265, 267:271, 287:288, 290:290, 292:293, 318:333, 335:335, 339:340, 345:345, 351:351. Redundancy feedback: This generated sequence is sufficiently non-redundant. Low reward. Significant improvements necessary.

\end{tcolorbox}
\end{figure*}

\begin{table*}[ht]
\centering
\setlength{\tabcolsep}{2pt} 
\renewcommand{\arraystretch}{1.2} 
\caption{Sequential revision vs Priority search, average \% improvement (or reduction) along with standard error and p-value from a paired t-test}
\label{table:stat:sig}
\begin{tabular}{l|ccc}
\toprule
\multicolumn{4}{l}{Single objective (pLDDT)} \\ \hline
& ss-pLDDT & TM-score & $C\alpha$-RMSD \\ 
Sequential Revision  & \; 19.4\% & \; 27.4\% & \; -15.0\% \\ 
Priority Search & \; 18.1 $\pm 0.85$\% & \; 56.9$\pm 4.92$\% & \; -31.2 $\pm 1.28$\% \\ 
& p=$1.61e^{-77}$ & p=$3.42e^{-57}$ & p=$8.71e^{-40}$ \\ \hline
\\ 
\multicolumn{4}{l}{Multi objective (ss-pLDDT, TM-score, $C\alpha$-RMSD)} \\ \hline
& ss-pLDDT & TM-score & $C\alpha$-RMSD \\ 
Sequential Revision  & 24.2$\pm 1.1$\% &  65.1$\pm 5.2$\% &  -32.6$\pm 1.3$\% \\ 
& p=$1.9e^{-81}$ & p=$2.1e{-65}$ & p=$1.42e^{-40}$ \\
Priority Search & 18.0 $\pm 0.82$\% \; & 67.7 $\pm 5.82$\% \; & -37.7 $\pm 1.24$\% \\
& p=$4.9e^{-81}$ & p=$6.1e^{-68}$ & p=$1.1e^{-45}$ \\
Cumulative & 38.5\% & 86.4\% & -44.4\% \\ \hline
\bottomrule
\end{tabular}
\end{table*}


\subsection{Characterizing the amino-acid substitution patterns}
\label{appendix:aminoacid:patterns}
We visualize the amino acid substitution matrix induced by LigandMPNN and our approach. For each method, we compute a residue-level substitution matrix that records the frequency with which an amino acid in the reference sequence is replaced by another amino acid in the designed sequence. Visualizing these matrices reveals systematic differences in substitution behavior: LigandMPNN exhibits more stereotyped, high-probability substitutions consistent with its learned inverse-folding priors, whereas RosettaSearch does more local exploration of the sequence space and relies on the overuse of structurally ``safe'' residues. Please see Figure \ref{fig:subsmatrix:ligandmpnn} and \ref{fig:native:as:starter} for details.

\subsection{Does the reasoning indicate hallucination?}
\label{appendix:hallucination}

We analyze the LLM expert opinion (from the Gemini-3 experts) to assess for evidence of hallucination in the reasoning produced by RosettaSearch, by verifying that the proposed sequence edits are actually applied and present in the generated sequences. 

Across the 22 cases deemed invalid by expert evaluation, failures overwhelmingly stem from reasoning–action inconsistencies rather than poor design intent. The dominant failure mode is mutation budget violation, where the LLM explicitly reasons about making at most 10 substitutions but produces sequences containing 15–30+ mutations. A second major class of errors involves hallucinated inputs or indexing mistakes, including incorrect assumptions about the native sequence, misidentified residue identities or positions, and reasoning about regions that do not correspond to the provided feedback.

Notably, a subset of failures also includes no-op generations, where the proposed sequence is unchanged despite detailed mutation plans in the reasoning.


\begin{figure*}[t]
        \centering
        \includegraphics[width=0.8\linewidth]{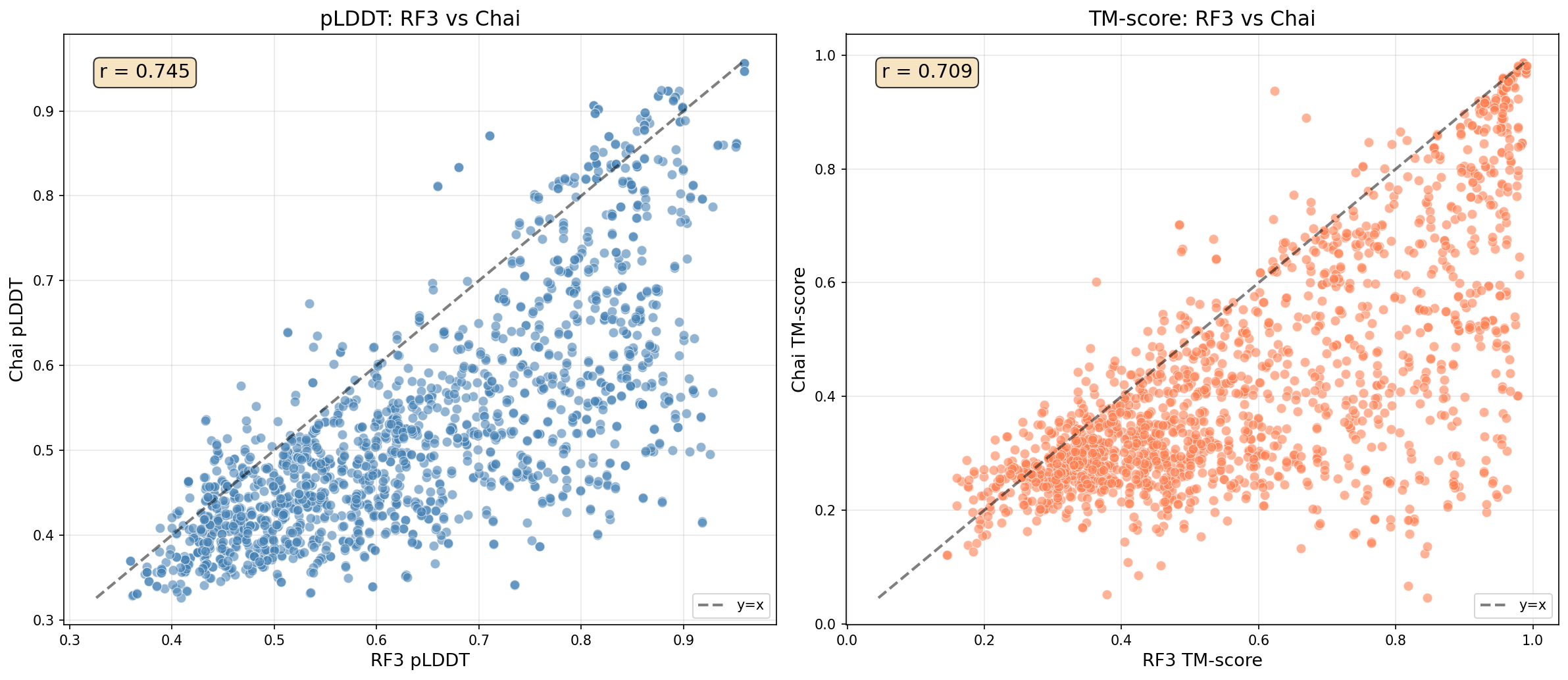}
        \caption{Comparison of the pLDDT and TM-score values of the predicted structures from RF3 and Chai-1, for sequences generated by the Search algorithm while doing multi-objective optimization. Each dot represents a protein sequence generated by RosettaSearch.}
    \label{fig:chai:vs:rf3}
\end{figure*}

\begin{figure}
    \centering
    \includegraphics[width=8cm]{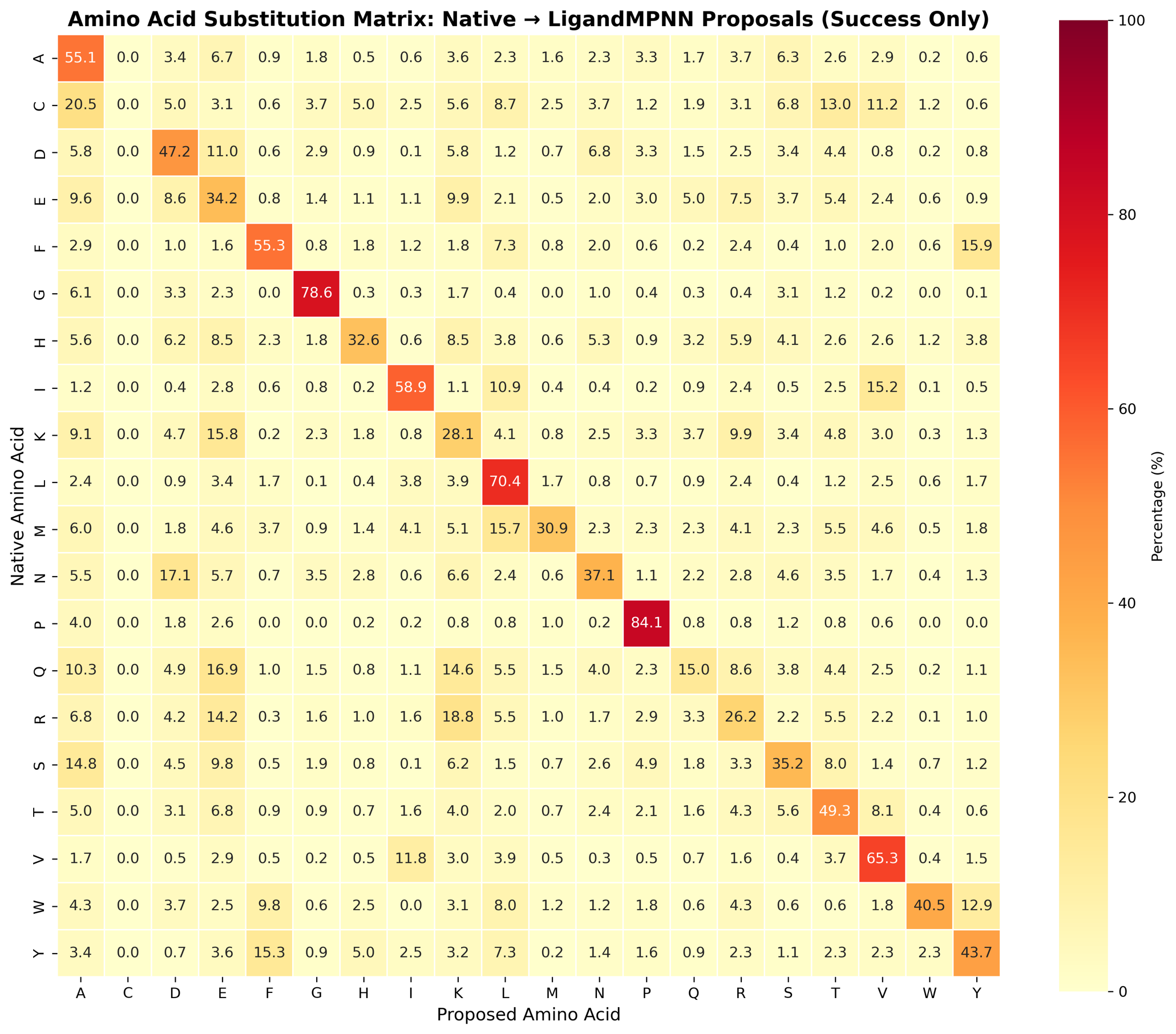}
    \caption{Substitution matrix from native protein sequences to the LigandMPNN designed sequences showing the biases of LigandMPNN.}
    \label{fig:subsmatrix:ligandmpnn}
\end{figure}

\begin{figure}
    \centering
    \includegraphics[width=8cm]{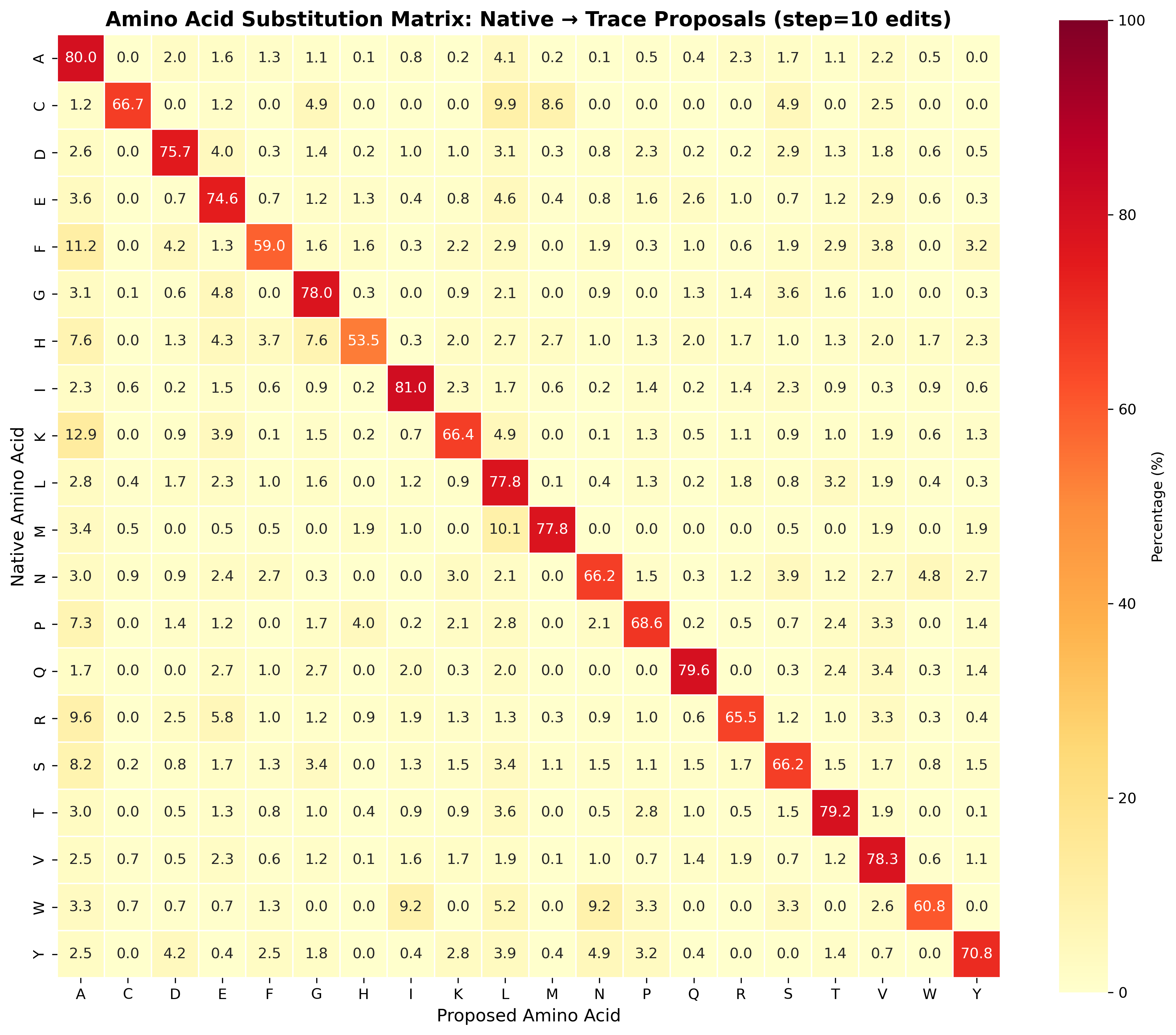}
    \caption{Substitution matrix from native protein sequences to the RosettaSearch designed sequences. This corresponds to setting (a) in our evaluation where the starting sequences are native sequences.}
    \label{fig:native:as:starter}
\end{figure}

\end{document}